\documentclass{llncs}

\usepackage{graphicx}
\usepackage{amsmath,amssymb}
\usepackage{mathtools}
\usepackage{xspace}
\usepackage{booktabs} 
\usepackage{xcolor}
\usepackage{colortbl}
\usepackage{multirow}
\usepackage{array}
\usepackage{hyperref}
\usepackage{url}
\usepackage{changepage}
\usepackage{todonotes} 

\usepackage{pdflscape}

\usepackage{pgfplots}
\pgfplotsset{compat=newest}

\pgfplotstableread[col sep=comma]{data/expectations.csv}\expectations

\usepackage{booktabs}
\usepackage{longtable}
\usepackage{nicefrac}
\usepackage{amssymb}
\usepackage{amsmath}
\usepackage{xspace}
\usepackage{mathtools}
\usepackage{diagbox}
\usepackage{graphics}
\usepackage{rotating}
\usepackage[algo2e,ruled,vlined,linesnumbered]{algorithm2e}
\newcommand{\assign}{\leftarrow}
\usepackage{graphicx}
\usepackage{subfigure}
\usepackage{ifthen}
\usepackage{wrapfig}
\usepackage{balance} 
\usepackage{epsfig}
\usepackage{xcolor, colortbl}
\usepackage[normalem]{ulem}

\graphicspath{{pictures/}}

\usepackage{array}
\newcolumntype{L}[1]{>{\raggedright\let\newline\\\arraybackslash\hspace{0pt}}m{#1}}
\newcolumntype{C}[1]{>{\centering\let\newline\\\arraybackslash\hspace{0pt}}m{#1}}
\newcolumntype{R}[1]{>{\raggedleft\let\newline\\\arraybackslash\hspace{0pt}}m{#1}}

\renewcommand{\epsilon}{\varepsilon}
\newcommand{\R}{\mathbb{R}}
\newcommand{\N}{\mathbb{N}}
\newcommand{\E}{\mathbb{E}}
\newcommand{\opl}{$(1+\lambda)$~EA\xspace}

\newcommand{\OM}{\textsc{OM}\xspace}
\newcommand{\onemax}{\textsc{OneMax}\xspace}

\DeclareMathOperator{\Bin}{Bin}

\DeclareMathOperator{\opt}{opt}

\DeclareMathOperator{\drift}{drift}
\DeclareMathOperator{\sbm}{sbm}
\newcommand{\shift}{0 \rightarrow 1}
\DeclareMathOperator{\select}{select}
\DeclareMathOperator{\flip}{\texttt{flip}}
\DeclareMathOperator{\RLS}{RLS}

\begin{document}

\title{Optimal Mutation Rates\texorpdfstring{\\}{} for the \texorpdfstring{$(1+\lambda)$}{(1+lambda)}~EA on OneMax}
\author{Maxim Buzdalov\inst{1} \and Carola Doerr\inst{2}}

\institute{ITMO University, Saint Petersburg, Russia,
\and Sorbonne Universit\'e, CNRS, LIP6, Paris, France}

\maketitle

\begin{abstract}
The OneMax problem, alternatively known as the Hamming distance problem, is often referred to as the ``drosophila of evolutionary computation (EC)'', because of its high relevance in theoretical and empirical analyses of EC approaches. It is therefore surprising that even for the simplest of all mutation-based algorithms, Randomized Local Search and the (1+1) EA, the optimal mutation rates were determined only very recently, in a GECCO 2019 poster.

In this work, we extend the analysis of optimal mutation rates to two variants of the $(1+\lambda)$~EA and to the $(1+\lambda)$~RLS. To do this, we use dynamic programming and, for the $(1+\lambda)$~EA, numeric optimization, both requiring $\Theta(n^3)$ time for problem dimension $n$. With this in hand, we compute for all population sizes $\lambda \in \{2^i \mid 0 \le i \le 18\}$ and for problem dimension $n \in \{1000, 2000, 5000\}$ which mutation rates minimize the expected running time and which ones maximize the expected progress. 

Our results do not only provide a lower bound against which we can measure common evolutionary approaches, but we also obtain insight into the structure of these optimal parameter choices. For example, we show that, for large population sizes, the best number of bits to flip is not monotone in the distance to the optimum.
We also observe that the expected remaining running time are not necessarily unimodal for the $(1+\lambda)$~EA$_{0 \rightarrow 1}$ with shifted mutation. 
\end{abstract}

\emergencystretch 3em 

\section{Introduction}
\label{sec:intro}

Evolutionary algorithms (EAs) are particularly useful for the optimization
of problems for which algorithms with proven performance guarantee are not known; e.g., due to a lack of knowledge, time, computational power, or access to problem data. 
It is therefore not surprising that we observe a considerable gap between the problems on which EAs are applied, and those for which rigorously proven analyses are available~\cite{DoerrN20}.  \\
If there is a single problem that stands out in the EA theory literature, this is the \onemax problem, which is considered to be ``the drosophila of evolutionary computation''~\cite{FialhoCSS09}. The \onemax problem asks to maximize the simple linear function that counts the number of ones in a bit string, i.e., $\OM(x)=\sum_{i=1}^n{x_i}$. This function is, of course, easily optimized by sampling the unique optimum $(1,\ldots,1)$. However, most EAs show identical performance on \onemax as on any problem asking to minimize the Hamming distance $H(z,\cdot)$ to an unknown string $z$, i.e., $f_z(x)=n-H(z,x)$, which is a classical problem studied in various fields of Computer Science, starting in the early 60s~\cite{erdos-renyi-two-problems}. In the analysis of EAs, \onemax typically plays the role of a benchmark problem that is easy to understand, and on which one can easily test the hill-climbing capabilities of the considered algorithm; very similar to the role of the sphere function in derivative-free numerical optimization~\cite{AugerH11,bbob-functions}.\\
Despite its popularity, 
and numerous deep results on the \onemax problem (see~\cite{DoerrN20} for examples), 
there are still a number of open questions, and this even for the simplest settings in which the problem is static and noise-free, and the algorithms under consideration can be described in a few lines of pseudo-code. 
One of these questions concerns the optimal mutation rates of the $(1+\lambda)$~EA, i.e., the algorithm which always keeps in memory a best-so-far solution $x$, and which samples in each iteration $\lambda$ ``offspring'' by applying standard bit mutation to $x$. By optimal mutation rates we refer to the values that minimize the expected optimization time, i.e., the average number of function evaluations needed until the algorithm evaluates for the first time an optimal solution.  
It is not very difficult to see that the optimal mutation rate of this algorithm as well as of its Randomized Local Search (RLS) analog 
(i.e., the algorithm applying a deterministic mutation strength rather than a randomly sampled one) 
depend only on the function value $\OM(x)$ of the current incumbent~\cite{Back92,BadkobehLS14,DoerrDY20}. 
However, even for $\lambda=1$ the optimal mutation rates were numerically computed only in the recent work~\cite{BuskulicD19}. 
Prior to~\cite{BuskulicD19}, only the rates that maximize the expected progress and those that yield asymptotically optimal running times (in terms of big-Oh notation) were known, see discussion below. \\
It was shown in~\cite{BuskulicD19} that the optimal mutation rates are not identical to those maximizing the expected progress, and that the differences can be significant when the current Hamming distance to the optimum is large. 
In terms of running time, however, the \emph{drift-maximizing} mutation rates are known to yield almost optimal performance, which is another result that was proven only recently~\cite{DoerrDY20} (more precisely, it was proven there for Randomized Local Search (RLS), but the result is likely to extend to the (1+1)~EA and its $(1+\lambda)$~variants).\\
\textbf{Our Contribution.} 
We extend in this work the results from~\cite{BuskulicD19} to the case $\lambda \in \{2^i \mid i \in [0..18] \}$. As in~\cite{BuskulicD19} we do not only focus on the standard $(1+\lambda)$~EA, but we also consider the $(1+\lambda)$ equivalent of RLS and we consider the $(1+\lambda)$~EA with the ``shift'' mutation operator suggested in~\cite{practice-aware}. The shift mutation operator $\shift$ flips exactly one randomly chosen bit when the sampled mutation strength of the standard bit mutation operator equals zero. \\
Differently from~\cite{BuskulicD19} we do not only store the optimal and the drift-maximizing parameter settings for the three different algorithms, but we also store the expected remaining running time of the algorithm that always applies the same fixed mutation rate as long as the incumbent has distance $d$ to the optimum and that applies the optimal mutation rate at all distances $d'<d$. With these values at hand, we can compute the \emph{regret} of each mutation rate, and summing these regrets for a given $(1+\lambda)$-type algorithm gives the exact expected running time, as well as the cumulative regret, which is the expected performance loss of the considered algorithm against the optimal strategy.  \\
Our results extend the main observation shared in~\cite{BuskulicD19}, which states that, for the (1+1)~EA, the drift-maximizing mutation rates are not always also optimal, to the $(1+\lambda)$~RLS and to both considered $(1+\lambda)$~EAs. We also show that the drift-maximizing and the optimal mutation rates are almost identical across different dimensions, when compared against the normalized distance $d/n$.\\ 
We also show that, for large population sizes, the optimal number of bits to flip is not monotone in the distance to the optimum. 
Moreover, we observe that the expected remaining running time is not necessarily unimodal for the $(1+\lambda)$~EA$_{0 \rightarrow 1}$ with shifted mutation. 
Another interesting finding is that some of the drift-maximizing mutation strengths of the $(1+\lambda)$~RLS with $\lambda>1$ are even, whereas it was proven in~\cite{DoerrDY20} that for the (1+1)~EA the drift-maximizing mutation strength must always be uneven. 
The distance $d$ at which we observe even drift-maximizing mutation strengths decreases with $\lambda$, whereas its frequency increases with~$\lambda$.\\ 
\textbf{Applications of Our Results in the Analysis of Parameter Control Mechanisms.} Apart from providing several data-driven conjectures about the formal relationship between the optimal and the drift-maximizing parameter settings of the investigated $(1+\lambda)$ algorithms, our results have immediate impact on the analysis of parameter control techniques. Not only do we provide an accurate lower bound against which we can measure the performance of other algorithms, but we can also very easily identify where potential performance losses originate from. We demonstrate such an example in Sec.~\ref{sec:app}, and recall here only that, despite its discussed simplicity, \onemax is a very commonly used test case for all types of parameter control mechanisms -- not only for theoretical studies~\cite{doerrs-parameter-control}, but also in purely empirical works~\cite{karafotias-overview,Thierens09}.\\ 
\textbf{OneMax Does Not Require Offspring Population.} It is well known that, for the optimization of \onemax, the $(1+1)$~EA is the most efficient among the $(1+\lambda)$~EAs~\cite{JansenJW05} when measuring performance by fitness evaluations. In practice, however, the $\lambda$~offspring can be evaluated in parallel, so that -- apart from mathematical curiosity -- the influence of the population size, the problem size, and the distance to the optimum on the optimal (and on the drift-maximizing) mutation rates also has practical relevance. \\
\textbf{Related Work.} 
Tight running time bounds for the $(1+\lambda)$~EA with \emph{static} mutation rate $p=c/n$ are proven in~\cite{GiessenW17}. For constant $\lambda$, these bounds were further refined in~\cite{GiessenW18}. The latter also presents optimal static mutation rates for selected combinations of population size $\lambda$ and problem size $n$. \\ 
For the here-considered \emph{dynamic} mutation rates, the following works are most relevant to ours. 
B\"ack~\cite{Back92} studied, by numerical means, the drift-maximizing mutation rates of the classic $(1+\lambda)$~EA with standard bit mutation, for problem size $n=100$ and for $\lambda \in \{1,5,10,20\}$. Mutation rates which minimize the expected optimization time in big-Oh terms were derived in~\cite[Theorem~4]{BadkobehLS14}. More precisely, it was shown there that the $(1+\lambda)$~EA using mutation rate 
$p(\lambda,n,d)=\max \{ 1/n, \ln(\lambda)/(n \ln(en/d))\}$ 
needs  
$O\left( \frac{n}{\ln \lambda} + \frac{n \log n}{\lambda}\right)$ 
function evaluations, on average, to find an optimal solution. This is asymptotically optimal among all $\lambda$-parallel mutation-only black-box algorithms~\cite[Theorem~3]{BadkobehLS14}. Self-adjusting and self-adaptive $(1+\lambda)$~EAs achieving this running time were presented in~\cite{doerrGWY-self-adjusting-mutation-rate} and~\cite{DoerrWY18}, respectively. 


\section{OneMax and $(1+\lambda)$ Mutation-Only Algorithms}
\label{sec:prelims}

As mentioned, the classical \onemax function $\OM$ simply counts the number of ones in the string, i.e., $\OM:\{0,1\}^n \to \R, x \mapsto \sum_{i=1}^n{x_i}$. For all algorithms discussed in this work, the behavior on \OM is identical to that on any of the problems 
$\OM_z:\{0,1\}^n \to \R, x \mapsto n-H(z,x):=|\{ i \in [n] \mid x_i \neq z_i\}|$. We study the maximization of these problems. 

\begin{algorithm2e}[t]%
  \small
	\textbf{Initialization:} 
	Sample $x \in \{0,1\}^{n}$ uniformly at random and evaluate $f(x)$\;
  \textbf{Optimization:}
	\For{$t=1,2,3,\ldots$}{
		\For{$i=1,\ldots,\lambda$}{
			\label{line:k}
			Sample $k(i) \sim D(n,f(x))$\;
			\label{line:mut} $y^{(i)} \assign \flip_{k(i)}(x)$\;
			evaluate $f(y^{(i)})$\;
			}
		$y \assign \select\left(\arg\max\{f(y^{(i)}) \mid i \in [\lambda]\} \right)$\; 	
		\lIf{$f(y)\geq f(x)$}{$x \assign y$\label{line:select}}	
	}
\caption{Blueprint of an elitist $(1+\lambda)$ unbiased black-box algorithm maximizing a function $f:\{0,1\}^n \to \R$. 
}
\label{alg:opl}
\end{algorithm2e}

Algorithm~\ref{alg:opl} summarizes the structure of the algorithms studied in this work. 
All algorithms start by sampling 
a uniformly chosen point $x$. 
In each iteration, $\lambda$ offspring $y^{(1)}, \ldots, y^{(\lambda)}$ are sampled from $x$, independently of each other. 
Each $y^{(i)}$ is created from the incumbent $x$ by flipping some $k(i)$ bits, which are pairwise different, independently and uniformly chosen (this is the operator $\flip$ in line~\ref{line:mut}). The best of these $\lambda$ offspring replaces the incumbent if it is at least as good as it (line~\ref{line:select}). When $\arg\max\{f(y^{(i)}) \mid i \in [\lambda]\}$ contains more than one point, the selection operator $\select$ chooses one of them, e.g., uniformly at random or via some other rule. As a consequence of the symmetry of \onemax, all results shown in this work apply regardless of the chosen tie-breaking rule. 

What is left to be specified is the distribution $D(n,f(x))$ from which the \emph{mutation strengths} $k(i)$ are chosen in line~\ref{line:k}. This is the only difference between the algorithms studied in this work.\\ 
\textbf{Deterministic vs. Random Sampling:} The Randomized Local Search variants (RLS) use a deterministic mutation strength $k(i)$, i.e., the distributions $D(n,f(x))$ are one-point Dirac distributions. We distinguish two EA variants: the one using standard bit mutation, denoted $(1+\lambda)$~EA$_{\sbm}$, and the one using the shift mutation suggested in~\cite{practice-aware}, which we refer to as $(1+\lambda)$~EA$_{\shift}$. 
	\emph{Standard bit mutation} uses the binomial distribution $\Bin(n,p)$ with $n$ trials and success probability $p$. The \emph{shift mutation} operator uses $\Bin_{0\rightarrow 1}(n,p)$, which differs from $\Bin(n,p)$ only in that all probability mass for $k=0$ is moved to $k=1$. That is, with shift mutation we are guaranteed to flip at least one bit, and the probability to flip exactly one bit equals $(1-p)^n+np(1-p)^{n-1}$. In both cases we refer to $p$ as the \emph{mutation rate}.  \\
	%
\textbf{Optimal vs. Drift-maximizing Rates:}	Our main interest is in the \emph{optimal} mutation rates, which minimize the expected time needed to optimize \onemax. 
		Much easier to compute than the optimal mutation rates are the \emph{drift-maximizing ones}, i.e., the values 
		which maximize the expected gain $\E[f(y)-f(x) \mid y \assign \flip_k(x), k \sim D(n,f(x))]$, see Sec.~\ref{sec:computation}. \\ 
\textbf{Notational Convention.} We omit the explicit mention of $(1+\lambda)$ when the value of $\lambda$ is clear from the context. 
Also, formally, we should distinguish between the \emph{mutation rate} (used by the EAs, see above) and the \emph{mutation strengths} (i.e., the number of bits that are flipped). However, to ease presentation, we will just speak of \textit{mutation rates} even when referring to the parameter setting for RLS.

\section{Computation of Optimal Parameter Configurations}
\label{sec:computation}

We compute the optimal parameters using the similar flavor of dynamic programming that has already been exploited in~\cite{BuskulicD19}.
Namely, as our algorithms behave on \OM\ identically regardless which parent they have at the particular distance to the optimum $d$, 
we compute the optimal parameters and the corresponding remaining time expectations for Hamming distance~$d$ to the optimum after we have computed them for all smaller distances $d'<d$. We denote by $T^{*}_{D,O}(n,\lambda,d)$ the minimal expected remaining time of a $(1+\lambda)$ algorithm with mutation rate distribution $D \in \{\RLS,\sbm,\shift\}$, optimality criterion $O \in \{\opt,\drift\}$, and population size $\lambda$ on a problem size $n \in \N$ when at distance $d \in [0..n]$. 
We also denote the distribution parameter (mutation strength in the case of RLS, mutation probability in the case of the EAs) 
by $\rho$, and the optimal distribution parameter for the current context as $\rho^{*}_{D,O}(n,\lambda,d)$.

Let $P_{n,D}(d, d', \rho)$ be the probability of sampling an offspring at distance $d'$ to the optimum, provided the parent is at distance $d$, the problem size is $n$, the distribution function is $D$, and the distribution parameter is $\rho$. 
The expected remaining time $T_{D,O}(n,\lambda,d,\rho)$, which assumes that at distance $d$ the algorithm consistently uses parameter $\rho$ and at all smaller distances it uses the optimal (time-minimizing or drift-maximizing, respectively) parameter for that distance, is then computed as follows:
\begin{align}
    T_{D,O}(n,\lambda,d,\rho) &= \frac{1}{(P_{n,D}(d, d, \rho))^{\lambda}} + \sum\nolimits_{d'=1}^{d-1} T^{*}_{D,O}(n,\lambda,d') \cdot P^{\lambda}_{n,D}(d, d',\rho), \label{eq:time-for-param}
\end{align}
where 		
$P^{\lambda}_{n,D}(d, d', \rho) = \left(\sum\nolimits_{t=d'}^{d} P_{n,D}(d,t,\rho)\right)^{\lambda} - \left(\sum\nolimits_{t=d'+1}^{d} P_{n,D}(d,t,\rho)\right)^{\lambda}.$


To compute $T^{*}_{D,O}(n,\lambda,d)$, Eq.~\eqref{eq:time-for-param} is used, where 
direct minimization of $\rho$ is performed when $O = \opt$, and the following drift-maximizing value of $\rho$ is substituted when $O = \drift$: 
$\rho_{n,D}(d) = \arg\max_{\rho} \sum_{d'=0}^{d-1} (d-d') \cdot P^{\lambda}_{n,D}(d, d',\rho)$. \\
%
Another difference to the work of~\cite{BuskulicD19} is in that we do not only compute the expected remaining running times 
$T^{*}_{D,O}(n,\lambda,d)$ when using the optimal mutation rates $\rho^{*}_{D,O}(n,\lambda,d)$, but we also compute and store
$T_{D,O}(n,\lambda,d,\rho)$ for suboptimal values of $\rho$. For $\RLS$ we do that for all possible values of $\rho$,
which are integers not exceeding $n$, while for the \opl we consider $\rho = 2^{i/5-10}/n$ for all $i \in [0;150]$.
We do this not only because it gives us additional insight into the sensitivity of $T_{D,O}(n,\lambda,d,\rho)$ with respect to $\rho$,
but it also offers a convenient way to detect deficits of parameter control mechanism; see Sec.~\ref{sec:app} for an illustrated example. 
Since our data base is hence much more detailed than that of~\cite{BuskulicD19}, we also re-consider the case $\lambda=1$. \\
Our code has the $\Theta(n^3)$ runtime and $\Theta(n^2)$ memory complexity. The code is available on GitHub~\cite{opl-code},
whereas the generated data is available on Zenodo~\cite{opl-data}.

\section{Optimal Mutation Rates and Optimal Running Times}
\label{sec:opt}

\begin{figure}[!t]
    \centering
    \includegraphics[width=\textwidth]{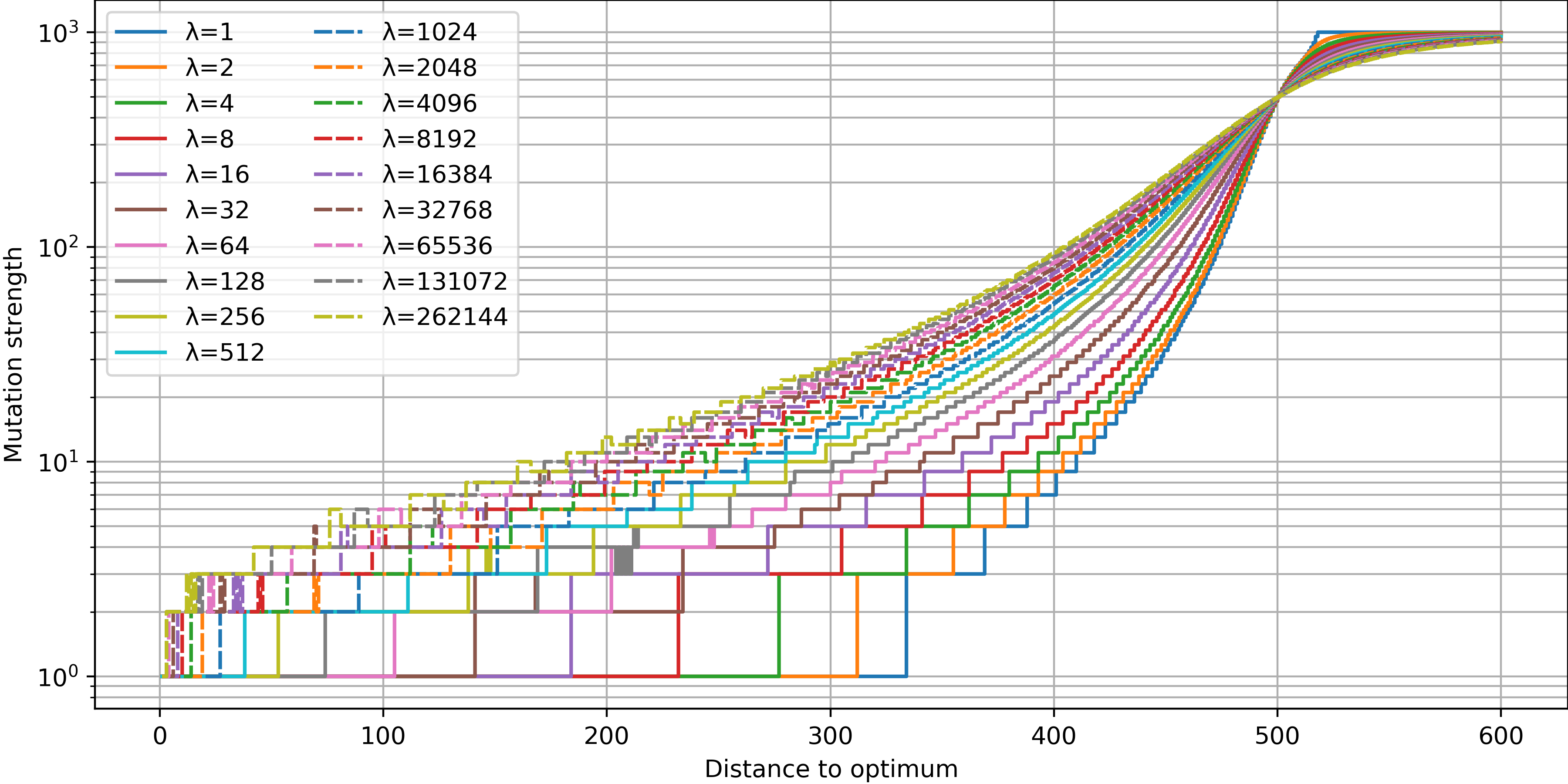}
    \caption{Optimal parameters $\rho^{*}_{\RLS,\opt}(n,\lambda,d)$ for different values of $\lambda$ and $n=1000$ as a function of $d$, the distance to the optimum}
    \label{fig:opt-RLS}
\end{figure}

Fig.~\ref{fig:opt-RLS} plots the optimal parameter settings $\rho^{*}_{\RLS,\opt}(n,\lambda,d)$ for fixed dimension $n=10^3$ and for different values of $\lambda$, in dependence of the Hamming distance $d$ to the optimum. We observe that the mutation strengths $\rho^{*}_{\RLS,\opt}(n,\lambda,d)$ are nearly monotonically increasing in $\lambda$, as a result of having more trials to generate an offspring with large fitness gain. 
We also see that, for some values of $\lambda$, the curves are not monotonically decreasing in $d$, but show small ``bumps''. Similar non-monotonic behavior can also be observed for drift-maximizing mutation strengths $\rho^{*}_{\RLS,\drift}(n,\lambda,d)$, as can be seen in Fig.~\ref{fig:bumps}. 

\begin{table}[!t]
\caption{Drifts for $n=30$, $\lambda=512$, $d=7,8$, $\rho\in[1..10]$.
See Appendix~\ref{sec:example-drift-derivation} for derivation}
\label{tbl:example-drift}
\setlength{\tabcolsep}{3pt}\centering
\begin{tabular}{c|*{10}{c}}
$d$ & $\rho=1$ & 2 & 3 & 4 & 5 & 6 & 7 & 8 & 9 & 10 \\\hline
7 & 0.5000 & 2.0000 & 2.9762 & 2.9604 & \textbf{3.0434} & 2.7009 & 2.5766 & 2.2292 & 1.7457 & 1.3854\\
8 & 0.5000 & 2.0000 & 2.9984 & \textbf{3.4601} & 3.3583 & 3.3737 & 3.2292 & 2.9124 & 2.7323 & 2.3445
\end{tabular}
\end{table}

\begin{figure}[!t]
    \centering
    \includegraphics[width=0.48\textwidth]{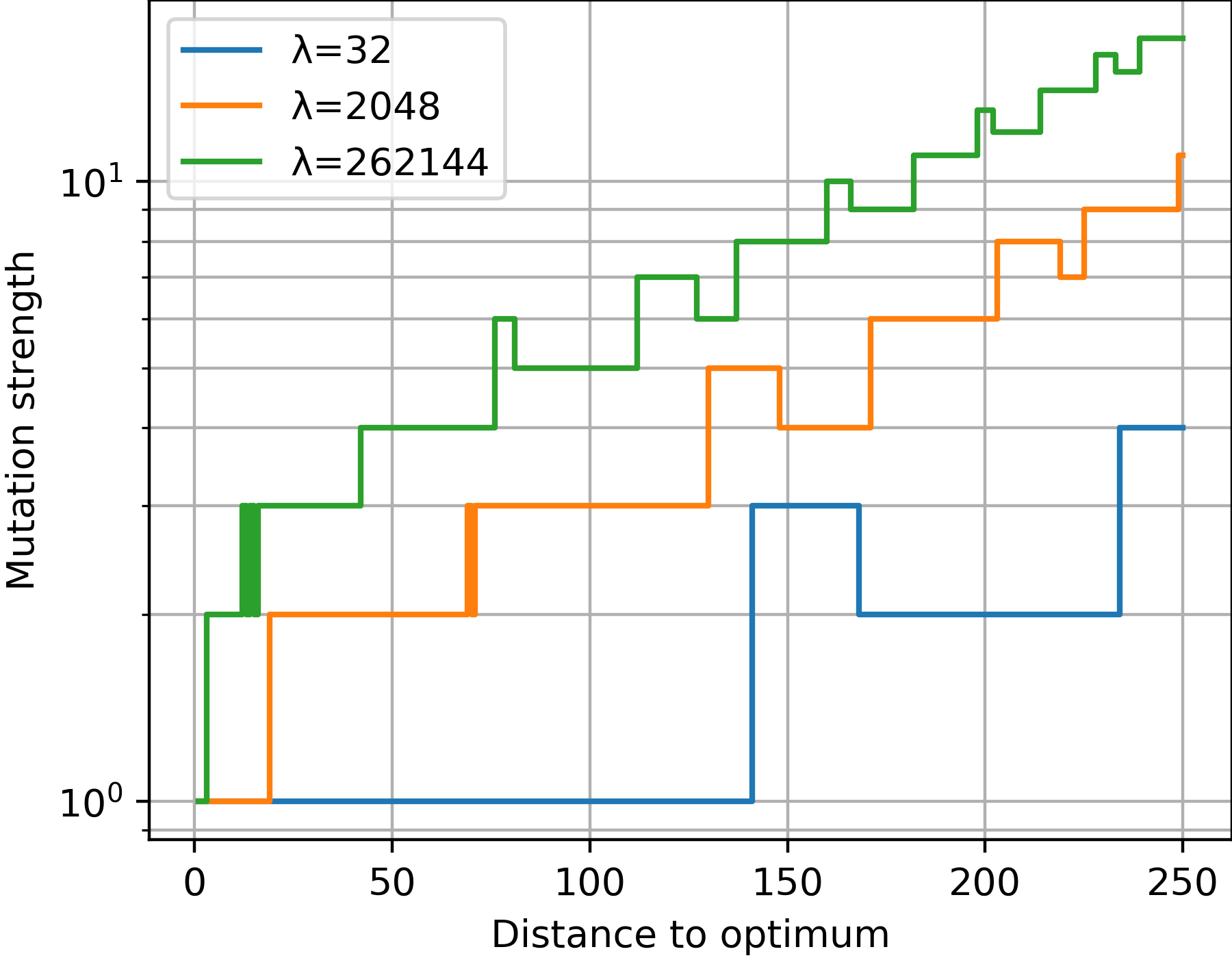}~~\includegraphics[width=0.48\textwidth]{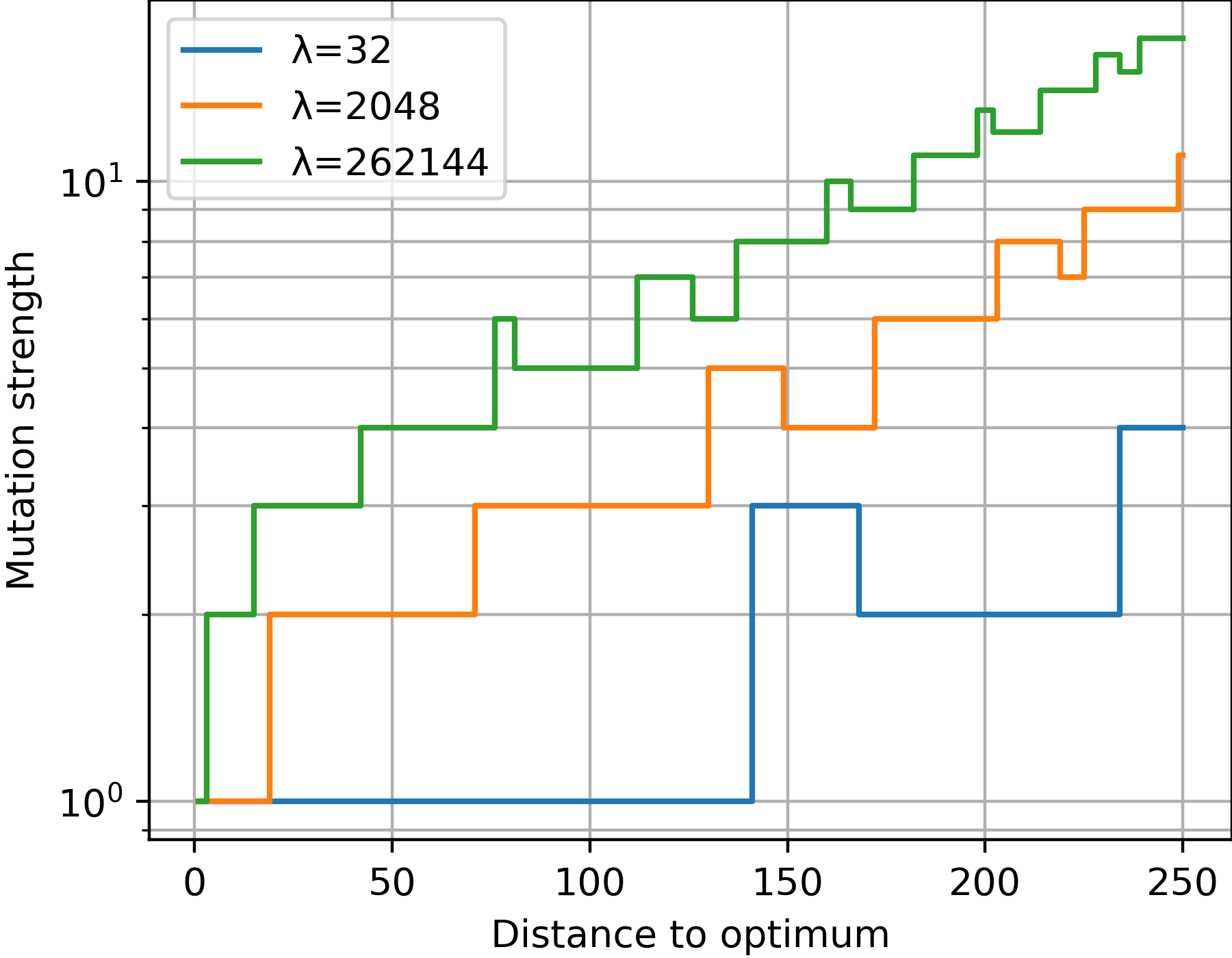}
    \caption{Non-monotonicity in optimal (left) and drift-optimal (right) mutation strengths for $n=1000$ and selected $\lambda$}
    \label{fig:bumps}
\end{figure}

We show now that these ``bumps'' are not just numeric precision artifacts, but rather a (quite surprising) feature of the parameter landscape. For a small example that can be computed by a human we consider $n=30$ and $\lambda = 512$. For $d=7$ and $8$, we compute the drifts for mutation strengths in $[1..10]$
(the details of these computation are given in Appendix~\ref{sec:example-drift-derivation}).
These values are summarized in Table~\ref{tbl:example-drift}. Here we see that the drift-maximizing mutation for $d=7$ is~$5$, whereas for $d=8$ it is~$4$. This example, in fact, serves two purposes: first, it shows that even the drift-maximizing strengths can be non-monotone, and second, that the drift-maximizing strengths can be even for non-trivial problem sizes, which -- as mentioned in the introduction -- cannot be the case when $\lambda=1$~\cite{DoerrDY20}.

\begin{figure}[!t]
    \centering
    \includegraphics[width=0.48\textwidth]{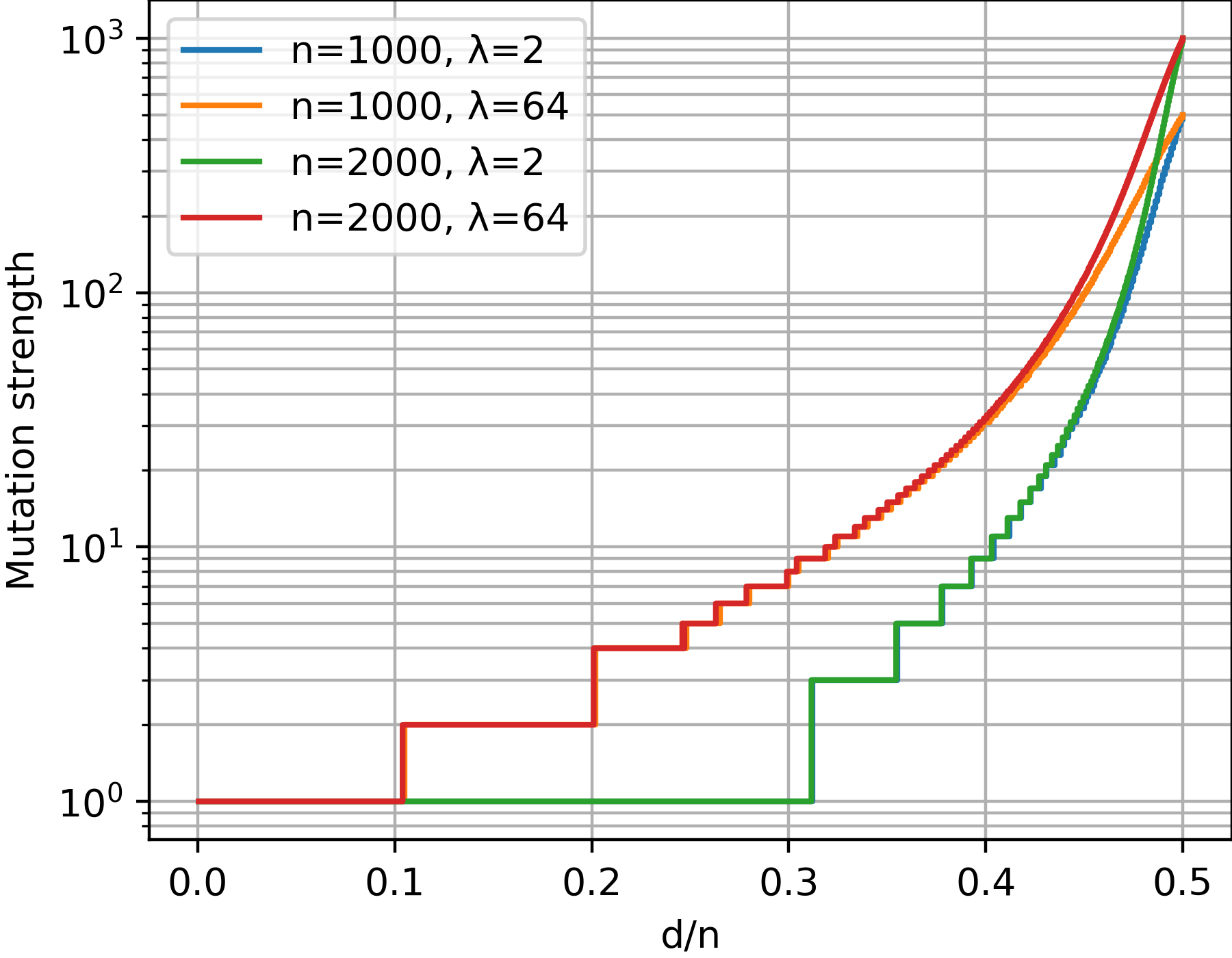}~~\includegraphics[width=0.48\textwidth]{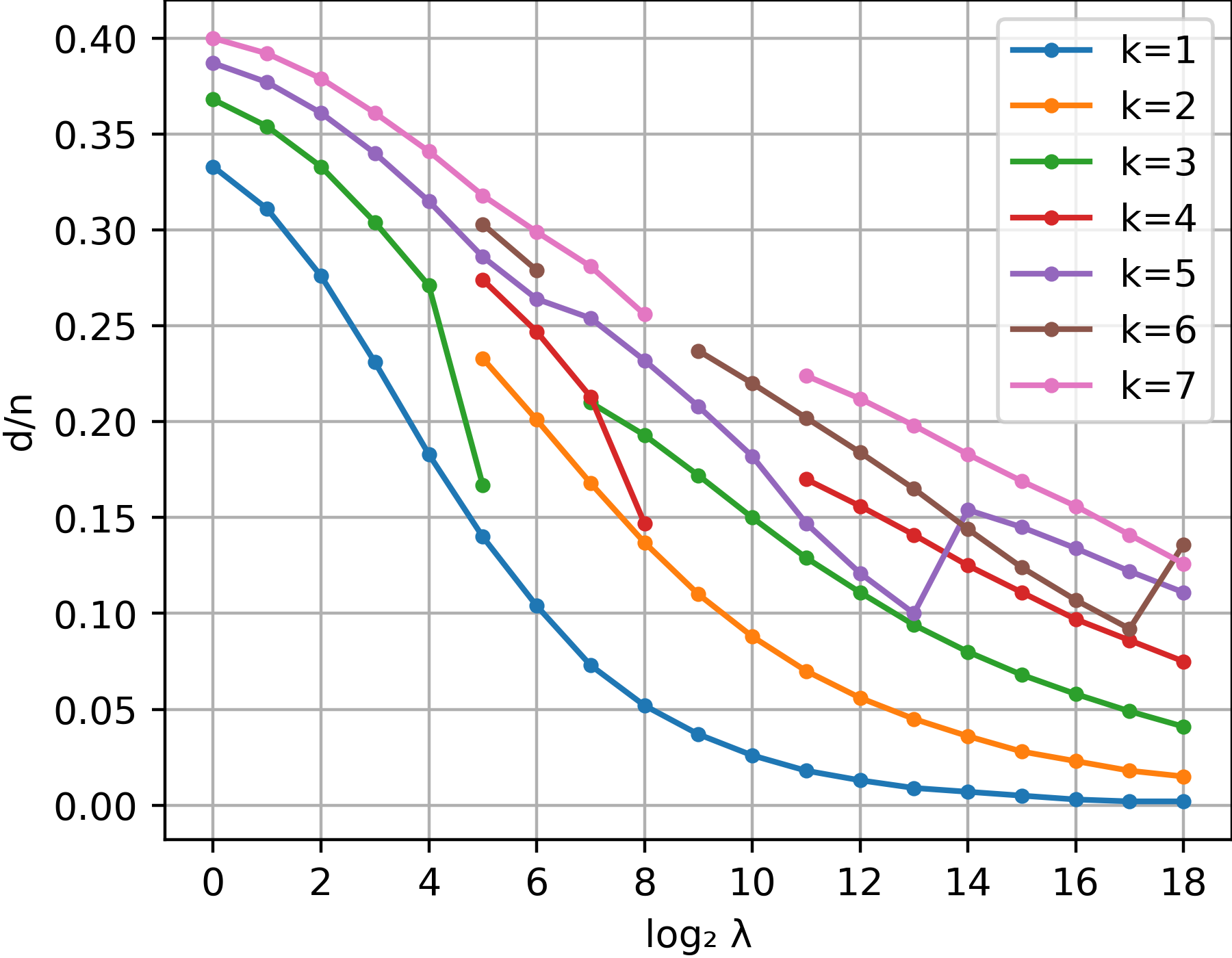}
    \caption{
		\textbf{Left:} $\rho^{*}_{\RLS,\opt}(n,\lambda,d)$ for $\lambda \in \{2,64\}$ and $n\in \{1k,2k\}$, in dependence of $d/n$.
		\textbf{Right:} Normalized maximal distance $d/n$ at which flipping $k\in[1..7]$ bits is optimal for $\RLS$, for $n=10^3$ and $\lambda\in \{2^i \mid 0 \le i \le 18\}$.
    }\label{fig:opt-RLS-64}
\end{figure}

In the left chart of Fig.~\ref{fig:opt-RLS-64} we show that at least small $\rho^{*}_{\RLS,\opt}(n,\lambda,d)$ are quite robust with respect to the problem dimension $n\in \{1,2\}\cdot 10^3$, if the Hamming distance $d$ to the optimum is appropriately scaled as $d/n$. The chart plots the curves for $\lambda \in \{2,64\}$ only, but the observation applies to all tested values of~$\lambda$. 
In accordance to our previous notes, we also see that for $\lambda=64$ there is a regime for which flipping two bits is optimal.
For small population sizes $\lambda$, we also obtain even numbers for certain regimes, but only for much larger distances. 

The maximal distances at which flipping $k$ bits is optimal are summarized in the chart on the right of Fig.~\ref{fig:opt-RLS-64}.
Note here that the curves are less smooth than one might have expected. For instance, for $n=10^3$, flipping three bits is never optimal for $\lambda=64$,
and flipping seven bits is never optimal for $\lambda = 2^9$ and $2^{10}$.


\begin{figure}[!t]
    \centering
    \includegraphics[width=0.48\textwidth]{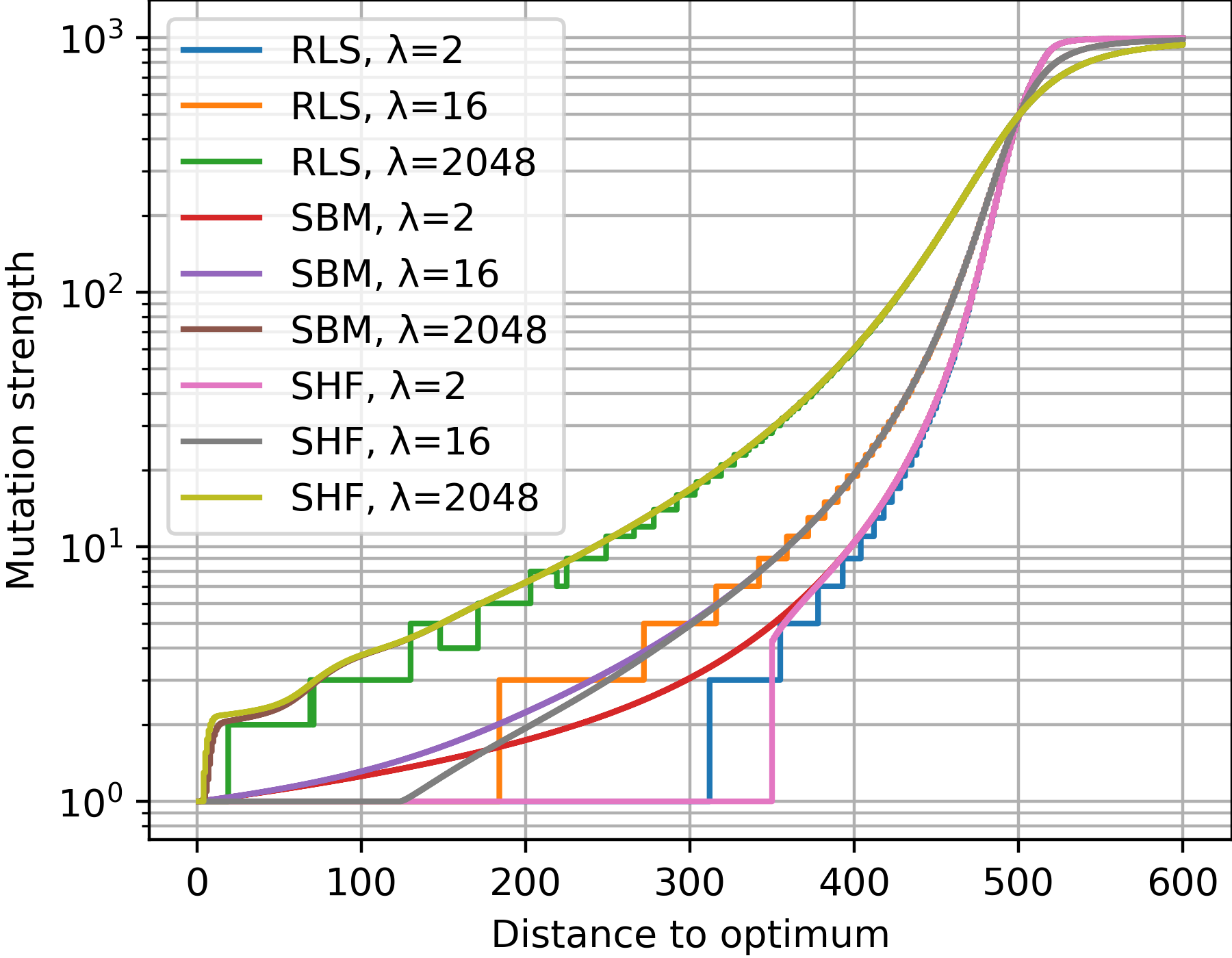}~~\includegraphics[width=0.48\textwidth]{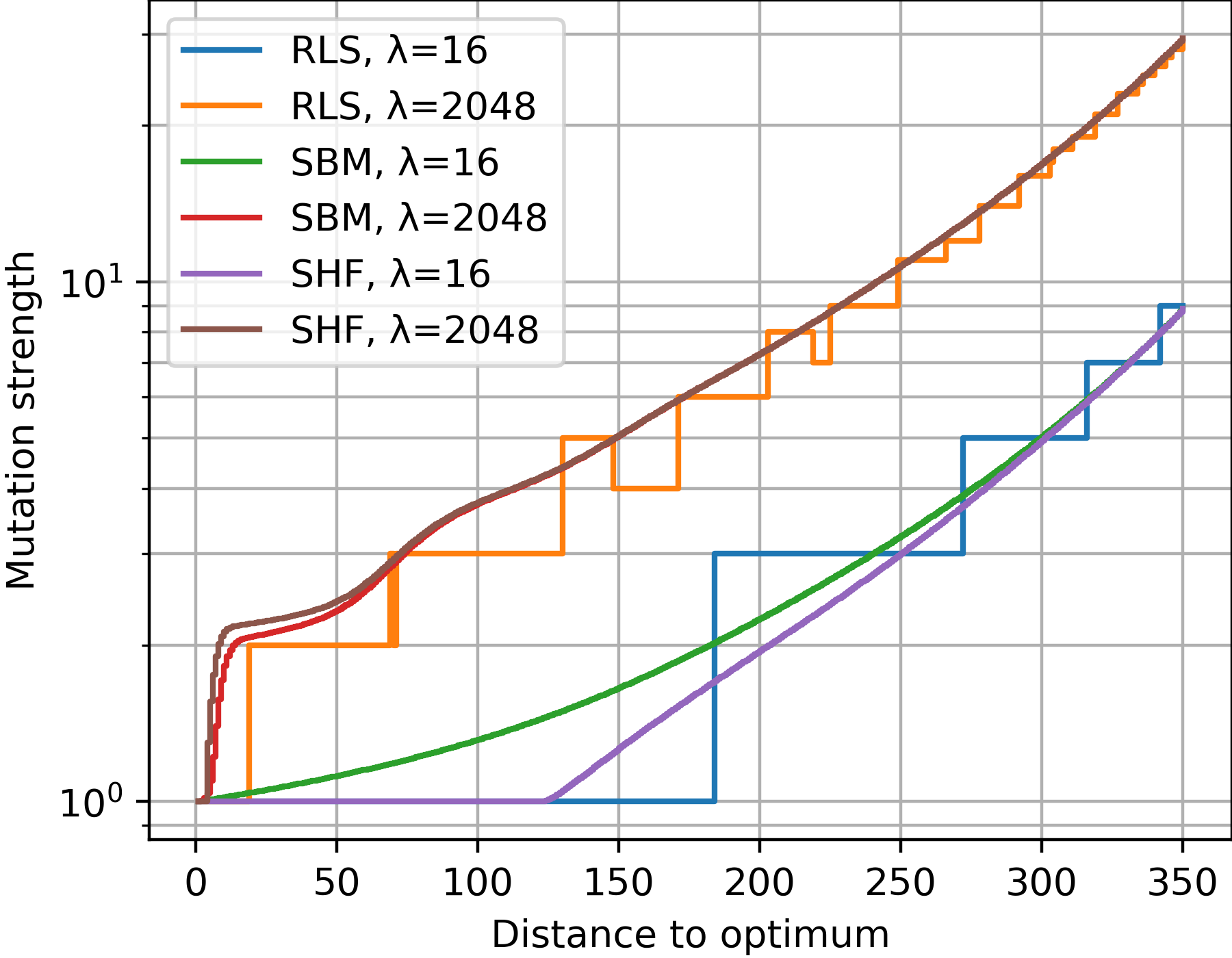}
    \caption{\textbf{Left:} Expected mutation strengths of the time-minimizing parameter settings for the $(1+\lambda)$~RLS and two $(1+\lambda)$~EAs, $\lambda\in\{2,16,2048\}$, using standard bit mutation (SBM) and shift mutation (SHF), respectively. Values are for $n=1000$ and plotted against the Hamming distance to the optimal solution. 
		\textbf{Right:} Same for $\lambda\in\{16, 2048\}$ with an emphasis on small distances.
	}
    \label{fig:compi}
\end{figure}

In Fig.~\ref{fig:compi} we compare the optimal (i.e., time-minimizing) parameter settings of the $(1+\lambda)$ variants of RLS, the EA$_{\shift}$, and the EA$_{\sbm}$. 
To obtain a proper comparison, we compare the mutation strength $\rho^{*}_{\RLS,\opt}(n,\lambda,d)$
with the expected number of bits that flip in the two EA variants, i.e., $n \rho^{*}_{\sbm,\opt}(n,\lambda,d)$
for the EA using standard bit mutation and $n \rho^{*}_{\sbm,\opt}(n,\lambda,d) + (1- \rho^{*}_{\sbm,\opt}(n,\lambda,d))^n$
for the EA using the shift mutation operator. We show here only values for $\lambda\in\{2,16,1024\}$, but the picture is similar for all evaluated $\lambda$.\\
We observe that, for each $\lambda$, the curves are close together. While for $\lambda=1$ the curves for standard bit mutation were always below that of RLS, we see here that this picture changes with increasing $\lambda$. We also see a sudden decrease in  the expected mutation strength of the shift operator when $\lambda$ is small. In fact, it is surprising to see that, for $\lambda=2$, the value drops from around 5.9 at distance 373 to 1 at distance 372. This is particularly interesting in light of a common tendency in state-of-the-art parameter control mechanisms to allow only for small parameter updates between two consecutive iterations. This is the case, for example, in the well-known one-fifth success rule~\cite{es-rechenberg,Devroye72,SchumerS68}. Parameter control techniques building on the family of reinforcement learning algorithms (see~\cite{FialhoCSS10} for examples) might catch such drastic changes more efficiently. \\
Non-surprisingly, the expected mutation strengths of the optimal standard bit mutation rate and the optimal shift mutation rate converge as the distance to the optimum increases. 


\section{Sensitivity of the Optimization Time w.r.t the Parameter Settings}
\label{sec:sensitivity}

In this section, we present our findings on the sensitivity of the considered $(1+\lambda)$ algorithms to their mutation parameters.
To do this, we use not only the expected remaining times $T^{*}_{D,O}(n,\lambda,d)$ that correspond to optimal parameter values,
but also $T_{D,O}(n,\lambda,d,\rho)$ for various parameter values $\rho$, which correspond to the situation when an algorithm uses the parameter $\rho$
while it remains at distance $d$, and switches to using the optimal parameter values (time-minimizing for $O=\opt$ and drift-maximizing for $O=\drift$, respectively) once the distance is improved. For reasons of space we focus on $O=\opt$.

We use distance-versus-parameter heatmaps as a means to show which parameter values are efficient.
As the non-optimality regret $\delta_{D,O}(n,\lambda,d,\rho) = T_{D,O}(n,\lambda,d,\rho) - T^{*}_{D,O}(n,\lambda,d)$ is asymptotically smaller than the remaining time,
we derive the color from the value $\tau(\rho) = \exp(-\delta_{D,O}(n,\lambda,d,\rho))$. Note that $\tau(\rho) \in (0;1]$, and the values close to one
represent parameters that are almost optimal by their effect. The parameters where $\tau(\rho) \approx 0.5$, on the other hand, correspond to a regret of roughly $0.7$,
that is, if the parameters satisfy $\tau(\rho) \ge 0.5$ throughout the entire optimization, the total expected running time is greater by at most $0.7n/2$ 
than the optimal time for this type of algorithms.

\begin{figure}[!t]
    \newcommand{\mywidth}{0.5\textwidth}
    \begin{tabular}{rr}
    \begin{tikzpicture}
        \begin{axis}[enlargelimits=false, axis on top, ylabel={$\rho$}, width=\mywidth, height=0.22\textheight]
            \addplot graphics [xmin=1, xmax=500, ymin=1, ymax=500, includegraphics={trim=0 0 500px 500px,clip}]{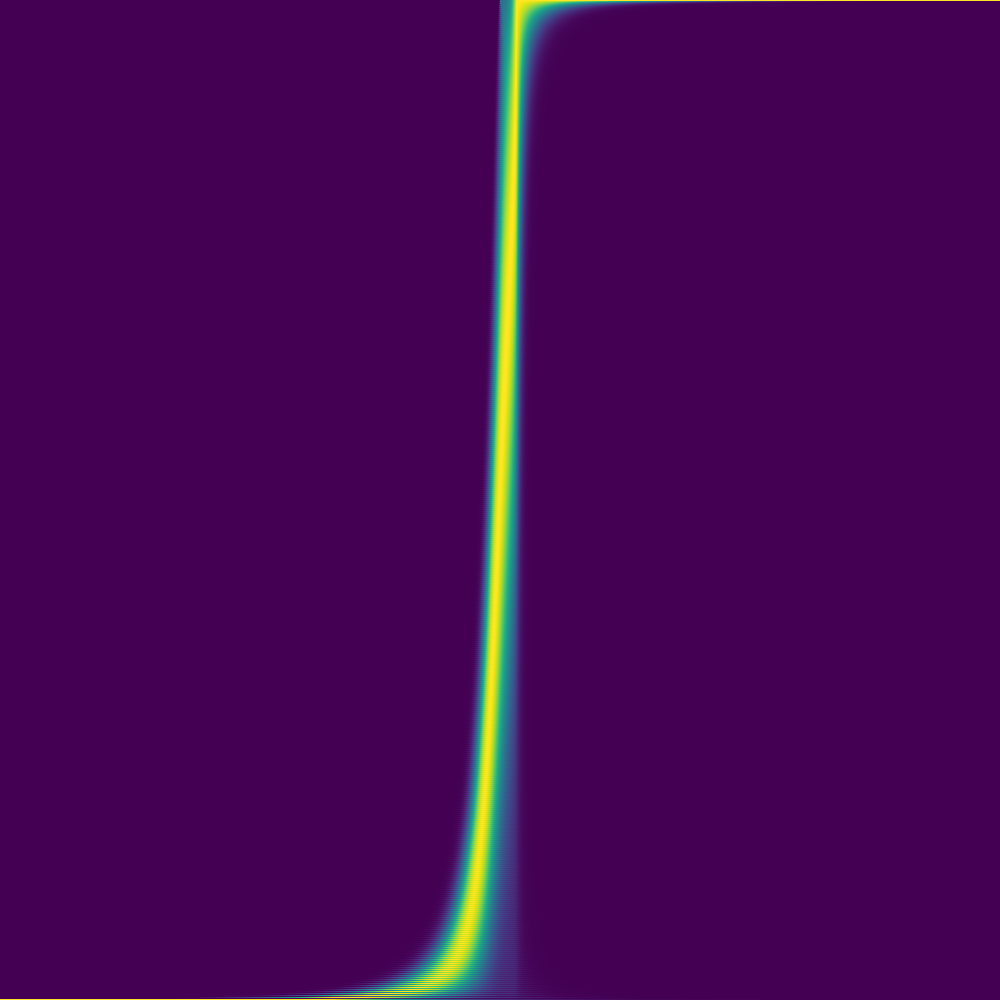}; 
        \end{axis}
    \end{tikzpicture} & \begin{tikzpicture}
        \begin{axis}[enlargelimits=false, axis on top, width=\mywidth, height=0.22\textheight, point meta min=0, point meta max=1, colorbar, colormap name=viridis, colorbar/width=2.5mm]
            \addplot graphics [xmin=1, xmax=500, ymin=1, ymax=500, includegraphics={trim=0 0 500px 500px,clip}]{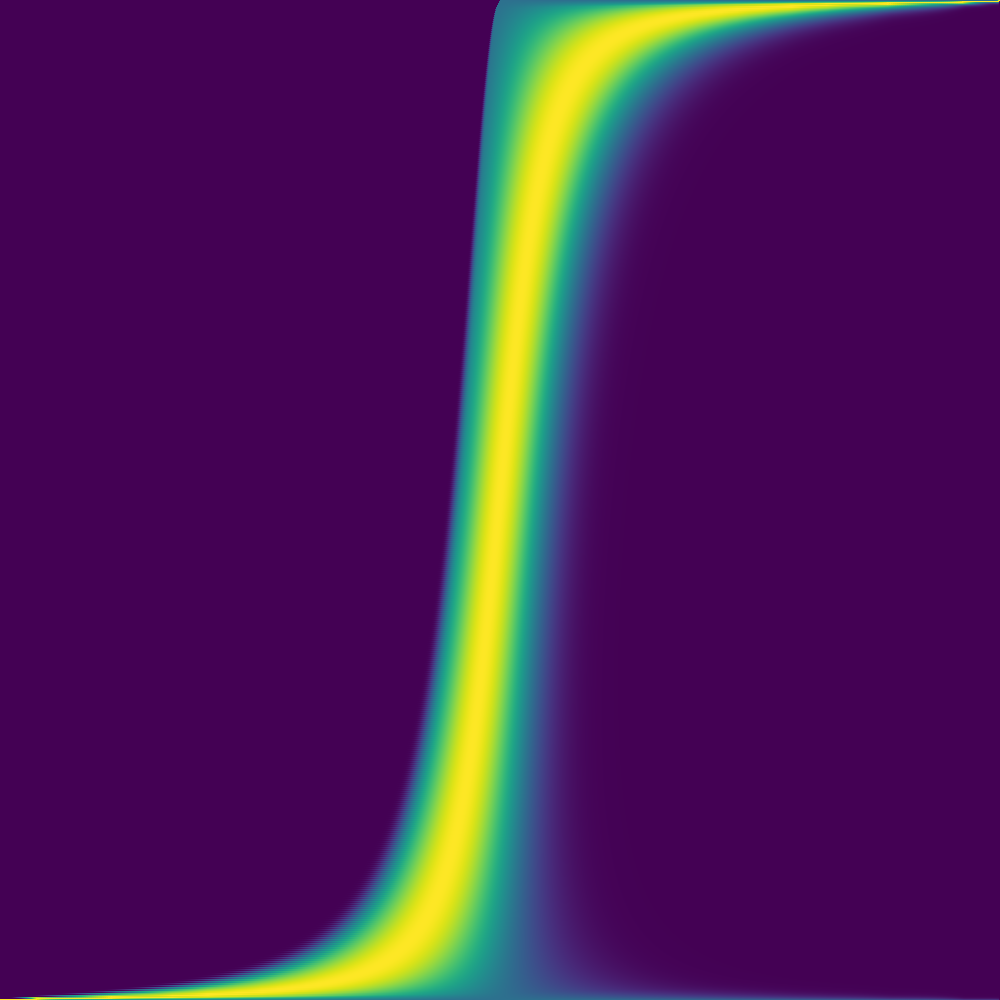}; 
        \end{axis}
    \end{tikzpicture} \\
    \begin{tikzpicture}
        \begin{axis}[enlargelimits=false, axis on top, xlabel={Distance to optimum}, ylabel={$\rho$}, width=\mywidth, height=0.2\textheight]
            \addplot graphics [xmin=1, xmax=500, ymin=1, ymax=50, includegraphics={trim=0 0 500px 950px,clip}]{optimal-rls-1000-1.png}; 
        \end{axis}
    \end{tikzpicture} & \begin{tikzpicture}
        \begin{axis}[enlargelimits=false, axis on top, xlabel={Distance to optimum}, width=\mywidth, height=0.2\textheight, point meta min=0, point meta max=1, colorbar, colormap name=viridis, colorbar/width=2.5mm]
            \addplot graphics [xmin=1, xmax=500, ymin=1, ymax=50, includegraphics={trim=0 0 500px 950px,clip}]{optimal-rls-1000-512.png}; 
        \end{axis}
    \end{tikzpicture}
    \end{tabular}
    \caption{Relative expected remaining optimization times for the $(1+\lambda)$~RLS$_{\opt}$ with parameters $n=10^3$, $\lambda=1$ (left) and $\lambda=512$ (right).
             The first row displays the general picture, the second row focuses on small mutation strengths}
    \label{fig:times-sensitivity-RLS}
\end{figure}
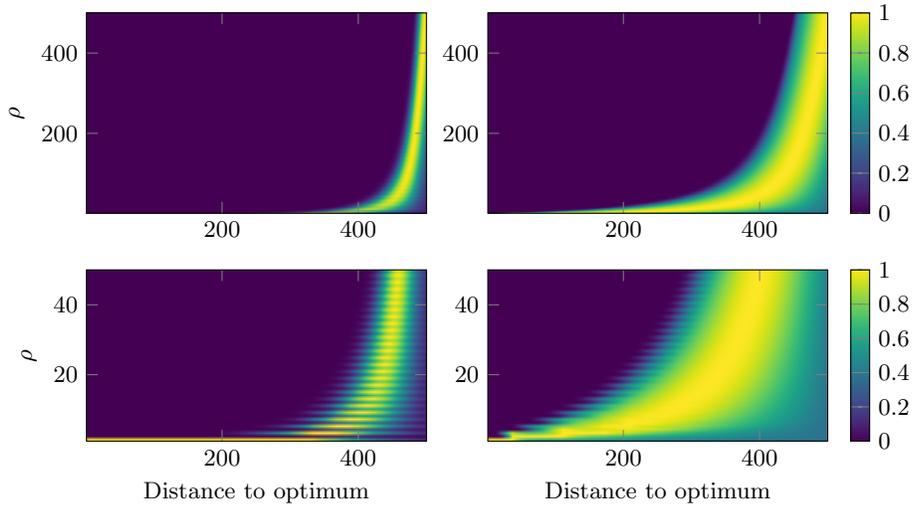

Fig.~\ref{fig:times-sensitivity-RLS} depicts these regrets for $\RLS_{\opt}$ on $n=10^3$ and $\lambda\in\{1, 512\}$. The stripes on the fine-grained plot for $\lambda=1$ expectedly indicate, as in~\cite{BuskulicD19}, that flipping an even number of bits is generally non-optimal when the distance to the optimum is small, which is the most pronounced for $\rho=2$. This also indicates that the parameter landscape of $\RLS$ is multimodal, posing another difficulty to parameter control methods. The parameter-time landscape remains multimodal for $\lambda=512$, but the picture is now much smoother around the optimal parameter values.

\begin{figure}[!t]
    \newcommand{\mywidth}{0.47\textwidth}
    \begin{tabular}{rr}
    \begin{tikzpicture}
        \begin{axis}[enlargelimits=false, axis on top, ylabel={$p\cdot n$}, width=\mywidth, height=0.22\textheight, ymode=log]
            \addplot graphics [xmin=1, xmax=500, ymin=9.765625E-4, ymax=512, includegraphics={trim=0 0 500px 0,clip}]{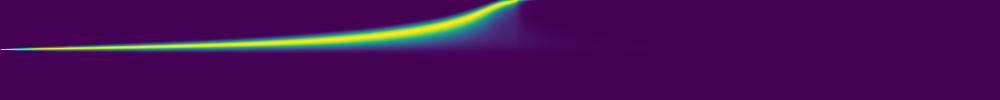}; 
        \end{axis}
    \end{tikzpicture} & \begin{tikzpicture}
        \begin{axis}[enlargelimits=false, axis on top, width=\mywidth, height=0.22\textheight, ymode=log, point meta min=0, point meta max=1, colorbar, colormap name=viridis, colorbar/width=2.5mm]
            \addplot graphics [xmin=1, xmax=500, ymin=9.765625E-4, ymax=512, includegraphics={trim=0 0 500px 0,clip}]{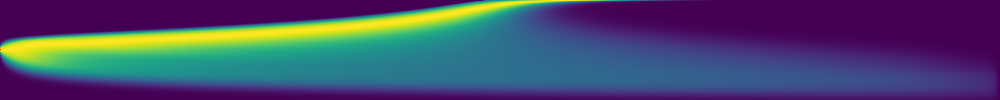}; 
        \end{axis}
    \end{tikzpicture} \\
    \begin{tikzpicture}
        \begin{axis}[enlargelimits=false, axis on top, xlabel={Distance to optimum}, ylabel={$p\cdot n$}, width=\mywidth, height=0.22\textheight, ymode=log]
            \addplot graphics [xmin=1, xmax=500, ymin=9.765625E-4, ymax=512, includegraphics={trim=0 0 500px 0,clip}]{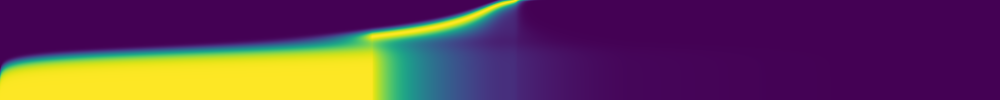}; 
        \end{axis}
    \end{tikzpicture} & \begin{tikzpicture}
        \begin{axis}[enlargelimits=false, axis on top, xlabel={Distance to optimum}, width=\mywidth, height=0.22\textheight, ymode=log, point meta min=0, point meta max=1, colorbar, colormap name=viridis, colorbar/width=2.5mm]
            \addplot graphics [xmin=1, xmax=500, ymin=9.765625E-4, ymax=512, includegraphics={trim=0 0 500px 0,clip}]{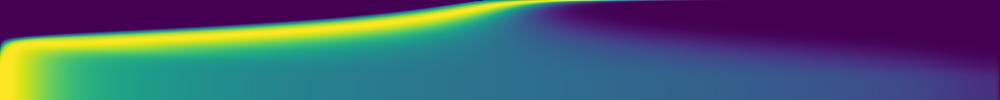}; 
        \end{axis}
    \end{tikzpicture}
    \end{tabular}

    \caption{Relative expected remaining optimization times for the $(1+\lambda)$~EA$_{\sbm,\opt}$ (top) and the $(1+\lambda)$~EA$_{\shift,\opt}$ (bottom)
             with $\lambda=1$ (left) and $\lambda=512$ (right)}
    \label{fig:times-sensitivity-shift}
\end{figure}

\begin{figure}[!t]
\begin{tikzpicture}
\begin{axis}[width=0.95\linewidth, height=0.22\textheight, xmode=log, legend pos=north west, xlabel={$\rho$}, ylabel={Time}]
\input{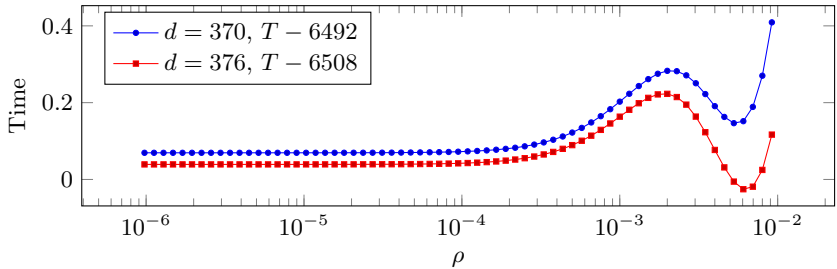}
\end{axis}
\end{tikzpicture}\par\vspace{-2ex}
\caption{Expected remaining optimization time of the $(1+\lambda)$~EA$_{\shift,\opt}$ as a function of the mutation probability~$\rho$}\label{fig:bimodality}
\end{figure}

Fig.~\ref{fig:times-sensitivity-shift} plots the regret for the $(1+\lambda)$~EA$_{\sbm}$ (top) and the $(1+\lambda)$~EA$_{\shift}$ (bottom) with $\lambda=1$ (left) and $\lambda=512$ (right).
The pictures for standard and shift mutations are very similar until the distance is so small that one-bit flips become nearly optimal. 
We also see that bigger population sizes result in a lower sensitivity of the expected remaining optimization time with respect to the mutation rate. 
In fact, we see that, even for standard bit mutation, parameter settings that are much smaller than the typically recommended mutation rates 
(e.g., $\rho=1/(10n)$) are also good enough when the distance is $\Omega(n)$, as the probability to flip at least one bit at least once is still quite large.\\
The plot for the $(1+1)$~EA$_{\shift}$ deserves separate attention. 
Unlike other plots in Fig.~\ref{fig:times-sensitivity-shift}, it demonstrates a bimodal behavior with
respect to the mutation probability $\rho$ even for quite large distances $d<n/2$. 
We zoom into this effect by displaying in Fig.~\ref{fig:bimodality} the expected remaining optimization times for $d\in\{370,376\}$.
Since mutation probability is a more likely candidate for parameter control than the number of bits to flip, this insight is even more important for parameter control.

\textbf{Drift-Maximization vs. Time-Minimization.} 
We note, without diving into the details, that the observation that the optimal mutation parameters are not identical to the drift maximizing ones, made in~\cite{BuskulicD19} for $(1+1)$~algorithms, extends to $(1+\lambda)$-type algorithms with $\lambda>1$. More precisely, it applies to all tested dimensions and population sizes $\lambda$. We note, though, that the disadvantage $T^{*}_{\RLS,\drift}(n,\lambda,d) - T^{*}_{\RLS,\opt}(n,\lambda,d)$ decreases with increasing $\lambda$. Since the difference is already quite small for the case $\lambda=1$ (e.g., for $n=1000$, it is 0.242), we conclude that this difference, albeit interesting from a mathematical perspective, has very limited relevance in empirical evaluations. This is good news for automated algorithm configuration techniques, as it implies that simple regret (e.g., in the terms of one-step progress) is sufficient to derive reasonable parameter values -- as opposed to requiring cumulative regret, which, as Sec.~\ref{sec:computation} shows, is much more difficult to track.
%
%

\section{Applications in Parameter Control}
\label{sec:app}

\begin{figure}[!t]
    \centering
    \begin{tikzpicture}
        \begin{axis}[width=0.9\textwidth, height=0.3\textheight, xmode=log, log base x=2, ymode=log, cycle list name=colorcycle, grid=both,
                     xlabel={Distance to optimum}, ylabel={Iterations}]
            \addplot+ coordinates {(2,4399.88)(4,2429.97)(8,1340.86)(16,809.82)(32,547.2)(64,402.89)(128,321.27)(256,275.41)(512,241.02)(1024,216.67)(2048,197.22)(4096,181.52)(8192,167.05)(16384,155.49)};
            \addlegendentry{\small $(1+\lambda)$ EA};
            \addplot+ coordinates {(2,9631.69)(4,4870.16)(8,2572.7)(16,1465.42)(32,841.01)(64,525.48)(128,350.51)(256,260.09)(512,211.24)(1024,177.72)(2048,154.49)(4096,136.56)(8192,124.36)(16384,112.57)};
            \addlegendentry{\small Two-rate, $p\ge\frac{1}{n}$};
            \addplot+ coordinates {(2,3825.23)(4,2041.04)(8,1155.44)(16,710.5)(32,484.33)(64,357.22)(128,283.69)(256,240.63)(512,204.27)(1024,179.61)(2048,157.08)(4096,140.34)(8192,127.1)(16384,115.55)};
            \addlegendentry{\small Two-rate, $p\ge\frac{1}{n^2}$};
            \addplot+ table[x=lambda, y=optimal rls] \expectations;
            \addlegendentry{\small Optimal $(1+\lambda) \RLS$};
            \addplot+ table[x=lambda, y=optimal shf] \expectations;
            \addlegendentry{\small Optimal $(1+\lambda) EA_{0\to1}$};
        \end{axis}
    \end{tikzpicture}
		\vspace{-10pt}
    \caption{Mean number of iterations of different $(1+\lambda)$~EAs vs. the expected number of iterations of $\RLS_{\opt}$ and $EA_{\opt,\shift}$ for $n=10^3$,
             as a function of the population size $\lambda$}
    \label{fig:runtimes}
\end{figure}
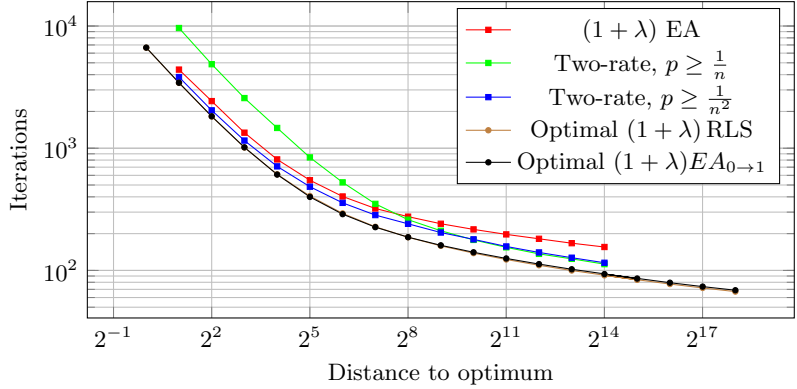

Fig.~\ref{fig:runtimes} displays the experimentally measured mean optimization times, averaged over 100 runs, of 
(1) the standard $(1+\lambda)$~EA with static mutation rate $\rho=1/n$,
(2) $\RLS_{\opt}$, 
(3) the $(1+\lambda)~EA_{\shift,\opt}$,  and 
of (4--5) the ``two-rate'' parameter control mechanism suggested in~\cite{doerrGWY-self-adjusting-mutation-rate}, superposed here to the $(1+\lambda)~\text{EA}_{\shift}$ with two different lower bounds $\rho_{\min}$ at which the mutation rate is capped. 

\pgfplotscreateplotcyclelist{colorcycle}{%
red,mark=square*,mark size=1\\
green,mark=square*,mark size=1\\
blue,mark=square*,mark size=1\\
brown,mark=*,mark size=1pt\\
black,mark=*,mark size=1pt\\
}

\pgfplotscreateplotcyclelist{traces}{%
red,mark=square*,mark size=0.5\\ 
red,mark=square*,mark size=0.5\\
red,mark=square*,mark size=0.5\\
red,mark=square*,mark size=0.5\\
black,mark=square*,mark size=0.5\\
black,mark=square*,mark size=0.5\\
black,mark=square*,mark size=0.5\\
}

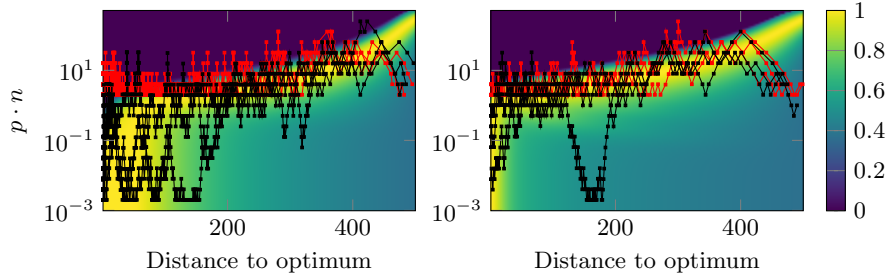
\begin{figure}[!t]
    \newcommand{\mywidth}{0.47\textwidth}
    \begin{tabular}{rr}
    \begin{tikzpicture}
        \begin{axis}[enlargelimits=false, axis on top, xlabel={Distance to optimum}, ylabel={$p\cdot n$}, width=\mywidth, height=0.22\textheight, ymode=log, cycle list name=traces]
            \addplot graphics [xmin=1, xmax=500, ymin=9.765625E-4, ymax=512, includegraphics={trim=0 0 500px 0,clip}]{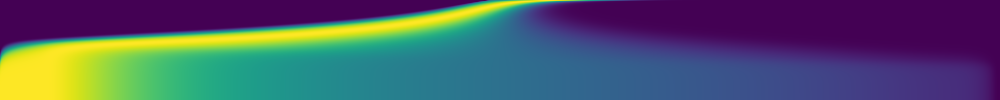}; 
            \addplot+ coordinates {(495,4.0)(488,8.0)(482,16.0)(470,32.0)(458,16.0)(447,32.0)(428,16.0)(413,32.0)(404,16.0)(400,32.0)(394,16.0)(386,32.0)(382,16.0)(373,32.0)(369,64.0)(365,128.0)(364,64.0)(359,128.0)(345,64.0)(342,32.0)(338,16.0)(334,32.0)(329,16.0)(327,32.0)(324,16.0)(321,32.0)(318,64.0)(318,32.0)(316,16.0)(313,8.0)(311,4.0)(309,2.0)(306,4.0)(303,8.0)(299,4.0)(296,8.0)(293,16.0)(291,8.0)(288,4.0)(286,2.0)(284,4.0)(282,8.0)(280,16.0)(279,32.0)(276,16.0)(271,8.0)(267,16.0)(266,8.0)(264,16.0)(262,32.0)(262,64.0)(262,32.0)(261,16.0)(261,8.0)(258,16.0)(256,8.0)(254,16.0)(254,8.0)(252,4.0)(251,2.0)(248,2.0)(247,2.0)(245,4.0)(244,2.0)(243,2.0)(241,4.0)(237,8.0)(236,4.0)(235,8.0)(233,4.0)(230,8.0)(227,4.0)(224,8.0)(223,16.0)(221,8.0)(219,16.0)(219,8.0)(217,16.0)(217,8.0)(216,4.0)(215,2.0)(214,2.0)(212,4.0)(211,8.0)(210,16.0)(210,32.0)(210,16.0)(209,32.0)(209,16.0)(209,32.0)(208,16.0)(208,8.0)(204,4.0)(203,2.0)(201,2.0)(199,4.0)(198,2.0)(197,4.0)(196,2.0)(195,2.0)(193,2.0)(192,2.0)(190,4.0)(189,2.0)(188,2.0)(187,2.0)(185,2.0)(184,2.0)(183,2.0)(182,2.0)(180,4.0)(179,8.0)(178,4.0)(175,2.0)(173,2.0)(170,2.0)(169,4.0)(168,8.0)(168,16.0)(168,32.0)(168,16.0)(168,32.0)(168,16.0)(168,8.0)(166,4.0)(164,2.0)(163,4.0)(162,8.0)(161,16.0)(161,8.0)(159,4.0)(157,2.0)(156,4.0)(155,8.0)(152,4.0)(151,2.0)(150,2.0)(149,4.0)(147,8.0)(147,4.0)(146,2.0)(144,2.0)(143,2.0)(141,4.0)(139,8.0)(138,4.0)(137,2.0)(136,2.0)(134,2.0)(133,2.0)(131,4.0)(130,2.0)(129,2.0)(128,4.0)(128,2.0)(127,4.0)(126,2.0)(125,2.0)(123,4.0)(122,2.0)(121,2.0)(120,2.0)(119,2.0)(118,2.0)(117,2.0)(116,2.0)(114,2.0)(113,2.0)(112,2.0)(111,4.0)(110,2.0)(109,2.0)(108,2.0)(108,2.0)(107,2.0)(106,4.0)(105,2.0)(104,2.0)(102,4.0)(101,2.0)(100,4.0)(99,2.0)(98,2.0)(97,4.0)(95,2.0)(94,2.0)(93,2.0)(92,2.0)(91,2.0)(90,2.0)(88,2.0)(87,4.0)(87,8.0)(86,4.0)(86,2.0)(85,4.0)(84,2.0)(83,4.0)(82,2.0)(80,2.0)(79,2.0)(79,2.0)(78,4.0)(77,2.0)(76,4.0)(75,2.0)(74,4.0)(74,2.0)(73,4.0)(72,2.0)(71,4.0)(70,2.0)(69,2.0)(68,2.0)(68,2.0)(68,4.0)(66,8.0)(66,4.0)(65,2.0)(64,4.0)(62,8.0)(62,4.0)(61,2.0)(60,2.0)(59,2.0)(58,2.0)(57,4.0)(57,8.0)(57,16.0)(57,8.0)(56,4.0)(56,2.0)(55,2.0)(54,2.0)(53,4.0)(52,2.0)(51,4.0)(51,2.0)(50,4.0)(50,2.0)(50,2.0)(49,4.0)(48,8.0)(47,4.0)(47,2.0)(46,4.0)(45,2.0)(45,2.0)(45,2.0)(45,2.0)(44,4.0)(43,2.0)(42,2.0)(42,2.0)(41,4.0)(41,2.0)(40,2.0)(39,2.0)(38,2.0)(37,4.0)(37,2.0)(36,4.0)(35,2.0)(34,2.0)(33,4.0)(33,2.0)(32,2.0)(32,2.0)(31,2.0)(30,2.0)(29,2.0)(29,4.0)(29,2.0)(29,2.0)(28,2.0)(27,2.0)(27,2.0)(27,2.0)(26,4.0)(26,8.0)(26,4.0)(26,8.0)(26,4.0)(25,2.0)(25,2.0)(25,4.0)(24,2.0)(23,2.0)(23,2.0)(23,4.0)(23,2.0)(23,2.0)(22,2.0)(22,2.0)(21,2.0)(20,2.0)(20,2.0)(19,4.0)(19,8.0)(19,4.0)(19,8.0)(19,16.0)(19,32.0)(19,16.0)(19,32.0)(19,16.0)(19,32.0)(19,16.0)(19,8.0)(19,4.0)(19,2.0)(19,4.0)(19,2.0)(19,4.0)(19,8.0)(19,4.0)(19,2.0)(19,4.0)(19,2.0)(18,4.0)(17,2.0)(17,4.0)(17,8.0)(17,16.0)(17,32.0)(17,16.0)(17,8.0)(17,16.0)(17,8.0)(17,16.0)(17,32.0)(17,16.0)(17,8.0)(17,16.0)(17,8.0)(17,4.0)(17,2.0)(17,4.0)(17,2.0)(17,2.0)(16,2.0)(16,4.0)(16,2.0)(16,2.0)(15,2.0)(14,4.0)(14,2.0)(13,2.0)(13,4.0)(12,2.0)(12,2.0)(12,2.0)(12,4.0)(12,2.0)(12,4.0)(12,2.0)(12,4.0)(11,2.0)(10,4.0)(10,2.0)(10,4.0)(10,2.0)(10,2.0)(10,2.0)(10,2.0)(9,4.0)(9,8.0)(9,4.0)(9,2.0)(9,2.0)(9,2.0)(9,2.0)(9,2.0)(8,4.0)(8,2.0)(8,2.0)(8,2.0)(8,2.0)(8,4.0)(8,2.0)(7,4.0)(7,2.0)(6,2.0)(6,2.0)(6,2.0)(6,4.0)(6,2.0)(6,4.0)(6,8.0)(6,4.0)(6,8.0)(6,4.0)(6,8.0)(6,4.0)(5,2.0)(5,2.0)(5,4.0)(5,8.0)(5,4.0)(5,8.0)(5,4.0)(5,2.0)(5,4.0)(5,2.0)(5,4.0)(5,2.0)(4,2.0)(4,2.0)(4,2.0)(4,2.0)(4,4.0)(4,2.0)(4,4.0)(4,2.0)(4,4.0)(4,8.0)(4,4.0)(4,2.0)(4,2.0)(4,2.0)(4,2.0)(4,2.0)(4,2.0)(4,2.0)(4,4.0)(4,2.0)(4,4.0)(4,2.0)(4,2.0)(4,2.0)(4,2.0)(4,4.0)(4,8.0)(4,4.0)(4,2.0)(4,2.0)(4,2.0)(4,4.0)(4,2.0)(4,2.0)(4,4.0)(4,2.0)(4,4.0)(4,2.0)(4,2.0)(4,2.0)(4,4.0)(4,2.0)(4,2.0)(4,4.0)(3,2.0)(3,2.0)(3,2.0)(3,4.0)(3,2.0)(3,4.0)(3,8.0)(3,16.0)(3,8.0)(3,4.0)(3,2.0)(2,4.0)(2,2.0)(2,2.0)(2,4.0)(2,2.0)(2,4.0)(2,2.0)(2,4.0)(2,2.0)(2,4.0)(2,2.0)(2,2.0)(2,2.0)(2,2.0)(2,2.0)(2,4.0)(2,2.0)(2,2.0)(2,4.0)(2,8.0)(2,4.0)(2,8.0)(2,16.0)(2,8.0)(2,4.0)(2,2.0)(2,2.0)(2,2.0)(2,2.0)(2,4.0)(2,8.0)(2,4.0)(2,2.0)(2,2.0)(2,2.0)(2,2.0)(1,2.0)(1,4.0)(1,2.0)(1,2.0)(1,2.0)(1,4.0)(1,2.0)(1,2.0)(1,2.0)(1,4.0)(1,2.0)(1,2.0)(1,2.0)(1,4.0)(1,2.0)(1,4.0)(1,8.0)(1,4.0)(1,2.0)(1,4.0)(1,2.0)(1,4.0)(1,2.0)(1,2.0)(1,4.0)(1,2.0)(1,4.0)(1,2.0)(1,4.0)(1,2.0)(1,4.0)(1,2.0)(1,4.0)(1,8.0)(1,4.0)(1,8.0)(1,4.0)(1,8.0)(1,4.0)(1,2.0)(1,2.0)(1,2.0)(1,4.0)(1,2.0)(1,2.0)(1,2.0)(1,2.0)(1,4.0)(1,2.0)(1,4.0)(1,2.0)(1,4.0)(1,2.0)(1,2.0)(1,4.0)(1,2.0)(1,2.0)(1,2.0)(1,2.0)(1,4.0)(1,8.0)(1,4.0)(1,2.0)(1,2.0)(1,2.0)(1,2.0)(1,4.0)(1,8.0)(1,16.0)(1,8.0)(1,4.0)(1,2.0)(1,4.0)(1,8.0)(1,4.0)(1,2.0)(1,2.0)(1,2.0)(1,2.0)(1,4.0)(1,2.0)(1,2.0)(1,2.0)};
            \addplot+ coordinates {(482,2.0)(478,4.0)(474,2.0)(470,4.0)(466,8.0)(458,16.0)(447,32.0)(431,64.0)(426,32.0)(421,16.0)(414,32.0)(406,64.0)(402,32.0)(399,16.0)(394,32.0)(388,16.0)(378,8.0)(372,16.0)(368,32.0)(366,64.0)(366,32.0)(361,16.0)(356,32.0)(350,16.0)(347,32.0)(345,16.0)(341,8.0)(336,16.0)(332,32.0)(328,16.0)(325,32.0)(322,16.0)(318,8.0)(314,16.0)(311,8.0)(308,4.0)(305,2.0)(303,2.0)(300,4.0)(295,8.0)(293,16.0)(289,32.0)(288,16.0)(283,32.0)(283,16.0)(281,32.0)(281,64.0)(281,128.0)(281,64.0)(281,128.0)(281,64.0)(281,32.0)(277,16.0)(275,32.0)(274,16.0)(272,32.0)(269,16.0)(266,8.0)(260,4.0)(256,2.0)(254,4.0)(252,8.0)(249,4.0)(248,2.0)(246,4.0)(244,2.0)(243,4.0)(241,8.0)(238,16.0)(234,8.0)(231,16.0)(230,32.0)(230,16.0)(229,8.0)(228,4.0)(226,2.0)(225,2.0)(223,2.0)(222,2.0)(221,2.0)(220,2.0)(218,2.0)(215,4.0)(213,2.0)(212,2.0)(210,2.0)(208,4.0)(205,8.0)(205,4.0)(202,8.0)(200,4.0)(198,2.0)(196,4.0)(194,2.0)(193,4.0)(192,8.0)(190,4.0)(189,2.0)(188,2.0)(187,4.0)(184,8.0)(183,16.0)(183,8.0)(180,4.0)(179,2.0)(178,2.0)(177,4.0)(175,8.0)(174,4.0)(173,8.0)(172,4.0)(171,2.0)(170,4.0)(169,8.0)(167,4.0)(166,2.0)(165,2.0)(164,2.0)(163,2.0)(162,2.0)(160,4.0)(157,8.0)(157,4.0)(156,2.0)(154,2.0)(153,4.0)(152,8.0)(151,4.0)(150,2.0)(148,4.0)(148,2.0)(147,4.0)(146,8.0)(145,16.0)(145,32.0)(145,16.0)(145,8.0)(145,4.0)(144,2.0)(143,2.0)(142,2.0)(141,2.0)(140,4.0)(139,2.0)(138,4.0)(137,2.0)(136,2.0)(135,4.0)(133,2.0)(132,2.0)(131,2.0)(130,4.0)(129,8.0)(128,4.0)(127,2.0)(126,2.0)(125,4.0)(124,2.0)(123,2.0)(122,2.0)(120,4.0)(118,2.0)(117,2.0)(116,4.0)(115,2.0)(114,2.0)(113,2.0)(113,2.0)(112,2.0)(110,2.0)(109,4.0)(108,8.0)(108,4.0)(107,2.0)(106,4.0)(105,2.0)(104,4.0)(103,2.0)(102,2.0)(101,2.0)(101,4.0)(100,8.0)(100,4.0)(99,2.0)(98,2.0)(98,2.0)(96,4.0)(95,2.0)(92,2.0)(91,2.0)(90,2.0)(88,2.0)(87,2.0)(86,4.0)(84,2.0)(83,2.0)(82,2.0)(81,2.0)(80,4.0)(79,8.0)(78,4.0)(77,2.0)(76,4.0)(75,2.0)(75,4.0)(75,2.0)(74,4.0)(73,8.0)(73,4.0)(72,8.0)(72,4.0)(71,2.0)(70,2.0)(69,4.0)(68,2.0)(67,2.0)(67,4.0)(66,8.0)(66,4.0)(65,2.0)(64,2.0)(63,2.0)(62,2.0)(61,2.0)(60,2.0)(59,2.0)(59,4.0)(58,2.0)(57,2.0)(56,2.0)(55,2.0)(54,2.0)(53,2.0)(52,2.0)(51,2.0)(51,4.0)(51,2.0)(51,2.0)(50,2.0)(49,4.0)(48,2.0)(48,2.0)(47,2.0)(46,2.0)(46,2.0)(45,2.0)(44,2.0)(43,2.0)(42,2.0)(41,2.0)(40,4.0)(39,2.0)(39,4.0)(39,2.0)(38,2.0)(38,2.0)(37,4.0)(37,2.0)(36,4.0)(36,2.0)(36,2.0)(35,2.0)(34,2.0)(34,2.0)(33,2.0)(32,4.0)(32,8.0)(32,4.0)(32,2.0)(32,2.0)(32,4.0)(31,2.0)(31,4.0)(31,2.0)(31,2.0)(30,4.0)(30,2.0)(30,4.0)(28,2.0)(28,4.0)(27,8.0)(27,4.0)(27,2.0)(26,4.0)(26,8.0)(26,4.0)(26,2.0)(25,2.0)(25,4.0)(25,2.0)(24,4.0)(24,8.0)(23,4.0)(22,8.0)(21,16.0)(21,8.0)(21,4.0)(21,2.0)(21,4.0)(21,2.0)(20,4.0)(19,2.0)(19,2.0)(19,4.0)(19,2.0)(18,2.0)(18,2.0)(18,4.0)(17,2.0)(17,2.0)(16,2.0)(16,2.0)(16,2.0)(16,2.0)(16,4.0)(16,2.0)(16,2.0)(16,4.0)(16,2.0)(16,2.0)(15,2.0)(15,4.0)(15,2.0)(15,2.0)(14,4.0)(14,2.0)(14,2.0)(13,2.0)(13,4.0)(13,2.0)(13,4.0)(13,2.0)(12,2.0)(12,2.0)(12,4.0)(12,2.0)(12,4.0)(12,2.0)(12,4.0)(12,2.0)(12,4.0)(12,2.0)(12,2.0)(12,2.0)(12,2.0)(12,2.0)(12,4.0)(12,2.0)(12,4.0)(12,2.0)(11,2.0)(11,4.0)(11,2.0)(10,4.0)(10,8.0)(10,4.0)(10,2.0)(10,2.0)(10,2.0)(10,2.0)(10,4.0)(10,2.0)(9,2.0)(9,4.0)(9,2.0)(9,2.0)(9,4.0)(9,2.0)(9,4.0)(9,2.0)(8,4.0)(8,2.0)(8,4.0)(8,2.0)(8,4.0)(8,8.0)(8,16.0)(8,8.0)(8,16.0)(8,8.0)(8,4.0)(8,2.0)(8,2.0)(8,2.0)(7,2.0)(7,2.0)(7,4.0)(7,2.0)(6,2.0)(6,4.0)(6,2.0)(6,4.0)(6,2.0)(6,4.0)(6,8.0)(6,4.0)(5,2.0)(5,2.0)(5,4.0)(5,2.0)(5,4.0)(5,2.0)(5,2.0)(4,4.0)(4,2.0)(4,2.0)(4,4.0)(4,2.0)(4,2.0)(4,2.0)(3,4.0)(3,2.0)(3,2.0)(3,2.0)(3,4.0)(3,2.0)(2,2.0)(2,4.0)(2,2.0)(2,4.0)(2,2.0)(2,2.0)(2,4.0)(2,2.0)(2,2.0)(2,2.0)(2,4.0)(2,8.0)(2,4.0)(2,8.0)(2,4.0)(2,8.0)(2,4.0)(2,2.0)(2,2.0)(2,4.0)(2,2.0)(2,2.0)(2,4.0)(2,2.0)(2,2.0)(2,2.0)(2,2.0)(2,4.0)(2,8.0)(1,4.0)(1,2.0)(1,2.0)(1,2.0)(1,4.0)(1,2.0)(1,4.0)(1,8.0)(1,4.0)(1,8.0)(1,16.0)(1,8.0)(1,16.0)(1,8.0)(1,4.0)(1,2.0)(1,4.0)(1,2.0)(1,2.0)(1,4.0)(1,2.0)(1,2.0)(1,4.0)(1,2.0)(1,2.0)(1,2.0)(1,4.0)(1,2.0)(1,2.0)(1,2.0)(1,2.0)(1,4.0)(1,2.0)(1,4.0)(1,8.0)(1,16.0)(1,8.0)(1,16.0)(1,8.0)(1,4.0)(1,8.0)(1,4.0)(1,2.0)(1,2.0)(1,2.0)(1,4.0)(1,2.0)(1,2.0)};
            \addplot+ coordinates {(463,4.0)(461,8.0)(452,16.0)(441,32.0)(433,16.0)(427,8.0)(421,16.0)(414,8.0)(410,16.0)(406,32.0)(401,64.0)(396,32.0)(393,64.0)(391,32.0)(387,64.0)(383,32.0)(378,16.0)(371,32.0)(369,64.0)(367,128.0)(367,64.0)(365,32.0)(363,64.0)(358,32.0)(355,64.0)(353,32.0)(347,16.0)(343,32.0)(339,64.0)(339,32.0)(337,16.0)(334,8.0)(330,16.0)(327,32.0)(326,16.0)(324,8.0)(321,4.0)(319,2.0)(315,2.0)(313,2.0)(310,4.0)(308,8.0)(307,4.0)(305,2.0)(302,2.0)(300,4.0)(297,8.0)(294,16.0)(292,32.0)(289,16.0)(288,32.0)(287,16.0)(285,8.0)(283,16.0)(282,8.0)(281,4.0)(280,8.0)(274,16.0)(273,8.0)(271,4.0)(269,2.0)(266,4.0)(264,2.0)(262,2.0)(261,4.0)(259,8.0)(257,4.0)(255,2.0)(251,2.0)(249,2.0)(247,4.0)(246,8.0)(245,4.0)(243,8.0)(240,4.0)(237,8.0)(231,16.0)(229,8.0)(227,16.0)(227,8.0)(224,4.0)(223,2.0)(220,2.0)(218,2.0)(215,2.0)(214,4.0)(212,2.0)(210,2.0)(209,2.0)(207,4.0)(205,2.0)(203,4.0)(201,8.0)(200,4.0)(198,8.0)(195,16.0)(195,32.0)(195,16.0)(195,8.0)(194,4.0)(192,8.0)(191,4.0)(190,2.0)(188,4.0)(187,2.0)(186,2.0)(184,4.0)(182,2.0)(181,2.0)(180,4.0)(179,2.0)(177,4.0)(176,2.0)(175,4.0)(174,8.0)(172,4.0)(171,2.0)(170,4.0)(169,2.0)(168,2.0)(167,2.0)(166,2.0)(165,2.0)(163,4.0)(162,2.0)(160,4.0)(158,2.0)(157,4.0)(156,2.0)(155,2.0)(154,4.0)(153,8.0)(151,4.0)(150,8.0)(149,4.0)(147,8.0)(146,4.0)(145,2.0)(144,2.0)(143,2.0)(142,2.0)(141,4.0)(140,2.0)(139,2.0)(138,4.0)(137,2.0)(136,2.0)(135,2.0)(134,4.0)(133,2.0)(130,2.0)(129,4.0)(128,2.0)(127,2.0)(125,4.0)(124,2.0)(123,2.0)(122,2.0)(121,2.0)(120,2.0)(118,2.0)(116,4.0)(115,2.0)(114,4.0)(112,2.0)(111,2.0)(110,4.0)(108,8.0)(107,4.0)(106,2.0)(105,4.0)(104,2.0)(101,4.0)(100,2.0)(99,2.0)(98,4.0)(97,2.0)(95,4.0)(94,2.0)(92,4.0)(91,2.0)(90,2.0)(89,4.0)(88,2.0)(87,2.0)(85,4.0)(84,2.0)(83,2.0)(80,4.0)(80,2.0)(79,2.0)(78,4.0)(77,2.0)(76,2.0)(75,4.0)(74,2.0)(73,2.0)(72,2.0)(71,2.0)(70,2.0)(69,2.0)(68,4.0)(68,2.0)(67,2.0)(66,2.0)(65,4.0)(65,8.0)(64,4.0)(64,2.0)(63,2.0)(62,2.0)(61,4.0)(61,2.0)(60,4.0)(60,8.0)(58,4.0)(58,2.0)(57,4.0)(57,2.0)(56,4.0)(56,2.0)(56,2.0)(55,2.0)(55,4.0)(54,2.0)(53,2.0)(52,2.0)(52,2.0)(51,2.0)(50,2.0)(50,2.0)(49,2.0)(48,2.0)(47,2.0)(46,4.0)(45,2.0)(44,2.0)(42,2.0)(41,2.0)(40,2.0)(39,2.0)(38,2.0)(37,4.0)(36,8.0)(36,4.0)(36,2.0)(36,2.0)(35,2.0)(34,2.0)(32,2.0)(31,2.0)(30,4.0)(30,2.0)(30,2.0)(30,4.0)(30,2.0)(29,4.0)(29,2.0)(29,4.0)(29,2.0)(28,4.0)(28,8.0)(28,16.0)(28,8.0)(28,4.0)(27,2.0)(27,4.0)(26,2.0)(26,4.0)(25,8.0)(25,4.0)(25,2.0)(25,2.0)(25,4.0)(25,2.0)(25,4.0)(25,2.0)(25,2.0)(24,2.0)(24,4.0)(24,2.0)(24,4.0)(23,2.0)(22,2.0)(21,2.0)(21,4.0)(20,8.0)(20,16.0)(20,8.0)(20,4.0)(20,2.0)(20,2.0)(19,2.0)(18,2.0)(18,2.0)(18,4.0)(18,8.0)(18,4.0)(18,2.0)(18,2.0)(18,2.0)(17,4.0)(17,8.0)(17,4.0)(17,8.0)(17,16.0)(17,8.0)(17,4.0)(16,2.0)(16,2.0)(15,4.0)(14,8.0)(14,16.0)(14,8.0)(14,4.0)(14,8.0)(14,16.0)(14,8.0)(14,16.0)(14,8.0)(14,4.0)(14,2.0)(14,4.0)(14,2.0)(13,2.0)(13,4.0)(13,2.0)(12,2.0)(12,4.0)(12,2.0)(12,2.0)(11,2.0)(11,2.0)(11,4.0)(11,2.0)(11,4.0)(10,2.0)(10,2.0)(10,2.0)(10,4.0)(9,2.0)(8,2.0)(8,2.0)(8,4.0)(8,2.0)(7,2.0)(7,4.0)(7,8.0)(7,4.0)(7,2.0)(7,4.0)(7,2.0)(6,2.0)(6,4.0)(5,2.0)(5,2.0)(5,4.0)(5,8.0)(5,16.0)(5,8.0)(5,16.0)(5,32.0)(5,16.0)(5,8.0)(5,4.0)(5,2.0)(5,4.0)(5,2.0)(4,4.0)(4,8.0)(4,4.0)(4,2.0)(4,2.0)(4,4.0)(4,2.0)(4,2.0)(4,2.0)(3,4.0)(3,2.0)(2,2.0)(2,2.0)(2,2.0)(2,4.0)(2,8.0)(2,4.0)(2,2.0)(2,2.0)(2,2.0)(1,2.0)(1,4.0)(1,2.0)(1,4.0)(1,8.0)(1,4.0)(1,2.0)(1,4.0)(1,2.0)(1,2.0)(1,2.0)(1,4.0)(1,2.0)(1,2.0)};
            \addplot+ coordinates {(497,16.0)(487,32.0)(474,64.0)(457,32.0)(445,64.0)(436,128.0)(423,250.0)(419,125.0)(412,250.0)(412,125.0)(404,62.5)(400,31.25)(396,15.625)(394,7.8125)(390,15.625)(383,31.25)(377,15.625)(373,31.25)(368,15.625)(365,31.25)(356,15.625)(353,7.8125)(350,15.625)(347,7.8125)(342,15.625)(335,7.8125)(332,3.90625)(330,1.953125)(327,3.90625)(324,7.8125)(322,15.625)(318,7.8125)(313,15.625)(312,7.8125)(307,3.90625)(302,7.8125)(301,15.625)(299,31.25)(297,15.625)(297,7.8125)(294,3.90625)(292,1.953125)(291,3.90625)(290,7.8125)(285,3.90625)(280,1.953125)(277,0.9765625)(275,1.953125)(273,3.90625)(270,1.953125)(268,3.90625)(267,1.953125)(265,3.90625)(263,7.8125)(262,15.625)(259,7.8125)(257,15.625)(255,7.8125)(253,15.625)(251,7.8125)(248,3.90625)(245,1.953125)(244,3.90625)(241,7.8125)(240,3.90625)(238,7.8125)(237,3.90625)(236,7.8125)(231,3.90625)(230,1.953125)(227,3.90625)(225,1.953125)(223,3.90625)(221,7.8125)(219,3.90625)(216,7.8125)(214,15.625)(214,7.8125)(212,15.625)(209,7.8125)(208,3.90625)(207,1.953125)(206,3.90625)(204,1.953125)(202,0.9765625)(201,1.953125)(199,3.90625)(197,1.953125)(196,3.90625)(191,1.953125)(189,0.9765625)(187,0.48828125)(185,0.9765625)(184,1.953125)(182,3.90625)(181,1.953125)(180,0.9765625)(179,1.953125)(178,0.9765625)(177,1.953125)(176,0.9765625)(174,1.953125)(173,3.90625)(170,7.8125)(167,3.90625)(165,1.953125)(164,0.9765625)(162,1.953125)(161,3.90625)(160,7.8125)(159,3.90625)(158,1.953125)(157,3.90625)(155,1.953125)(154,3.90625)(152,1.953125)(151,3.90625)(150,1.953125)(149,3.90625)(147,1.953125)(143,3.90625)(142,1.953125)(141,0.9765625)(140,1.953125)(139,3.90625)(138,1.953125)(135,3.90625)(133,1.953125)(132,3.90625)(130,7.8125)(129,15.625)(129,7.8125)(129,3.90625)(128,7.8125)(127,3.90625)(126,1.953125)(125,0.9765625)(124,1.953125)(123,0.9765625)(122,1.953125)(121,3.90625)(119,1.953125)(118,0.9765625)(117,1.953125)(114,3.90625)(113,1.953125)(112,0.9765625)(110,1.953125)(109,3.90625)(108,1.953125)(107,3.90625)(106,7.8125)(104,15.625)(104,31.25)(104,15.625)(104,7.8125)(104,3.90625)(103,1.953125)(102,3.90625)(101,1.953125)(100,3.90625)(99,1.953125)(98,0.9765625)(97,0.48828125)(96,0.9765625)(95,0.48828125)(94,0.244140625)(93,0.48828125)(92,0.9765625)(91,0.48828125)(90,0.9765625)(89,0.48828125)(88,0.244140625)(87,0.48828125)(86,0.244140625)(85,0.1220703125)(84,0.244140625)(83,0.48828125)(82,0.9765625)(81,1.953125)(80,0.9765625)(78,1.953125)(77,0.9765625)(76,1.953125)(75,0.9765625)(75,0.48828125)(74,0.244140625)(73,0.1220703125)(72,0.244140625)(71,0.48828125)(70,0.9765625)(69,0.48828125)(68,0.244140625)(67,0.1220703125)(66,0.06103515625)(65,0.1220703125)(64,0.06103515625)(63,0.1220703125)(62,0.06103515625)(61,0.1220703125)(60,0.244140625)(59,0.48828125)(58,0.244140625)(57,0.48828125)(56,0.244140625)(55,0.1220703125)(54,0.06103515625)(53,0.1220703125)(52,0.06103515625)(51,0.030517578125)(51,0.06103515625)(50,0.030517578125)(49,0.0152587890625)(48,0.030517578125)(47,0.0152587890625)(46,0.00762939453125)(45,0.0152587890625)(44,0.030517578125)(43,0.0152587890625)(42,0.00762939453125)(41,0.0152587890625)(40,0.00762939453125)(39,0.003814697265625)(38,0.00762939453125)(37,0.0152587890625)(37,0.00762939453125)(36,0.0152587890625)(35,0.030517578125)(34,0.06103515625)(34,0.1220703125)(33,0.06103515625)(32,0.030517578125)(31,0.0152587890625)(30,0.00762939453125)(29,0.003814697265625)(28,0.00762939453125)(27,0.0152587890625)(26,0.030517578125)(25,0.06103515625)(24,0.1220703125)(23,0.244140625)(22,0.1220703125)(22,0.06103515625)(21,0.1220703125)(20,0.244140625)(19,0.1220703125)(18,0.244140625)(17,0.1220703125)(16,0.244140625)(15,0.48828125)(15,0.9765625)(15,1.953125)(15,3.90625)(15,1.953125)(15,0.9765625)(14,0.48828125)(14,0.9765625)(14,0.48828125)(13,0.244140625)(12,0.48828125)(11,0.244140625)(11,0.48828125)(11,0.9765625)(11,0.48828125)(11,0.244140625)(10,0.1220703125)(9,0.244140625)(8,0.1220703125)(7,0.244140625)(7,0.48828125)(7,0.244140625)(7,0.1220703125)(7,0.244140625)(7,0.48828125)(6,0.244140625)(6,0.48828125)(6,0.244140625)(6,0.1220703125)(5,0.06103515625)(4,0.1220703125)(4,0.244140625)(3,0.48828125)(2,0.244140625)(2,0.48828125)(2,0.9765625)(2,1.953125)(2,0.9765625)(2,1.953125)(2,3.90625)(2,1.953125)(2,0.9765625)(2,0.48828125)(2,0.9765625)(2,1.953125)(2,0.9765625)(2,1.953125)(2,0.9765625)(2,1.953125)(2,0.9765625)(2,0.48828125)(2,0.244140625)(2,0.48828125)(2,0.9765625)(2,0.48828125)(2,0.244140625)(2,0.1220703125)(2,0.06103515625)(2,0.030517578125)(2,0.06103515625)(2,0.1220703125)(1,0.06103515625)(1,0.030517578125)(1,0.0152587890625)(1,0.030517578125)(1,0.0152587890625)(1,0.00762939453125)(1,0.0152587890625)(1,0.030517578125)(1,0.0152587890625)(1,0.00762939453125)(1,0.003814697265625)(1,0.00762939453125)(1,0.003814697265625)(1,0.00762939453125)(1,0.0152587890625)};
            \addplot+ coordinates {(479,4.0)(475,2.0)(472,4.0)(467,8.0)(460,16.0)(448,32.0)(439,16.0)(430,8.0)(419,16.0)(414,8.0)(411,16.0)(403,8.0)(400,16.0)(396,32.0)(388,64.0)(386,32.0)(382,16.0)(377,8.0)(372,16.0)(367,8.0)(364,4.0)(357,8.0)(354,4.0)(351,2.0)(347,4.0)(344,2.0)(342,4.0)(338,8.0)(336,4.0)(333,2.0)(330,1.0)(328,0.5)(327,1.0)(325,0.5)(324,0.25)(322,0.125)(321,0.25)(320,0.125)(319,0.0625)(318,0.125)(317,0.25)(316,0.5)(314,1.0)(313,2.0)(310,4.0)(307,2.0)(305,1.0)(304,2.0)(301,4.0)(297,8.0)(294,16.0)(288,8.0)(286,4.0)(283,8.0)(280,16.0)(278,8.0)(277,4.0)(275,8.0)(275,4.0)(273,8.0)(271,16.0)(268,8.0)(267,4.0)(264,8.0)(261,4.0)(259,2.0)(257,1.0)(255,0.5)(253,1.0)(251,0.5)(249,1.0)(247,2.0)(245,4.0)(243,2.0)(241,4.0)(238,2.0)(237,4.0)(235,2.0)(234,1.0)(233,0.5)(232,1.0)(231,2.0)(228,4.0)(226,2.0)(224,1.0)(222,0.5)(221,0.25)(220,0.5)(219,1.0)(218,2.0)(217,1.0)(216,2.0)(214,4.0)(211,8.0)(210,4.0)(209,2.0)(208,1.0)(207,2.0)(206,4.0)(204,2.0)(203,4.0)(200,2.0)(199,1.0)(198,2.0)(197,1.0)(195,2.0)(194,1.0)(192,0.5)(191,0.25)(190,0.5)(188,0.25)(187,0.5)(186,0.25)(185,0.125)(184,0.0625)(183,0.125)(182,0.25)(181,0.125)(180,0.0625)(179,0.125)(177,0.0625)(176,0.03125)(175,0.0625)(174,0.03125)(173,0.0625)(172,0.03125)(171,0.0625)(170,0.03125)(169,0.015625)(168,0.03125)(167,0.015625)(166,0.03125)(165,0.0625)(164,0.125)(163,0.0625)(162,0.03125)(161,0.0625)(160,0.03125)(159,0.0625)(158,0.125)(157,0.25)(156,0.5)(155,1.0)(154,2.0)(153,4.0)(152,2.0)(151,4.0)(150,2.0)(149,4.0)(148,2.0)(147,1.0)(145,2.0)(145,1.0)(144,0.5)(143,1.0)(141,2.0)(140,1.0)(139,0.5)(138,0.25)(137,0.125)(136,0.25)(135,0.5)(134,0.25)(133,0.125)(132,0.25)(131,0.5)(130,0.25)(129,0.5)(128,0.25)(127,0.5)(126,0.25)(125,0.5)(124,1.0)(122,2.0)(121,1.0)(120,0.5)(119,1.0)(118,0.5)(117,0.25)(116,0.5)(115,1.0)(114,2.0)(113,1.0)(112,2.0)(111,1.0)(110,0.5)(109,0.25)(108,0.125)(107,0.0625)(106,0.03125)(105,0.0625)(104,0.03125)(103,0.015625)(102,0.0078125)(101,0.015625)(100,0.0078125)(99,0.015625)(98,0.0078125)(97,0.015625)(96,0.03125)(95,0.015625)(94,0.0078125)(93,0.015625)(92,0.0078125)(91,0.00390625)(90,0.002)(89,0.004)(88,0.008)(87,0.016)(86,0.008)(85,0.004)(84,0.002)(83,0.004)(82,0.002)(81,0.004)(80,0.002)(79,0.004)(78,0.008)(78,0.016)(77,0.032)(76,0.016)(75,0.008)(74,0.016)(73,0.032)(72,0.016)(71,0.032)(70,0.064)(69,0.128)(68,0.064)(67,0.128)(67,0.064)(66,0.128)(65,0.256)(64,0.512)(63,1.024)(62,2.048)(61,1.024)(60,2.048)(60,1.024)(60,2.048)(60,4.096)(59,8.192)(58,16.384)(58,8.192)(58,16.384)(58,32.768)(58,16.384)(58,8.192)(57,16.384)(57,32.768)(57,16.384)(57,8.192)(57,16.384)(57,8.192)(56,4.096)(56,8.192)(56,4.096)(54,2.048)(53,1.024)(53,2.048)(52,4.096)(52,8.192)(52,4.096)(52,2.048)(51,1.024)(50,0.512)(49,1.024)(48,2.048)(48,1.024)(47,2.048)(46,1.024)(45,2.048)(44,1.024)(43,0.512)(42,1.024)(41,2.048)(41,4.096)(40,8.192)(40,4.096)(40,2.048)(40,4.096)(40,8.192)(40,4.096)(39,8.192)(39,16.384)(39,32.768)(39,16.384)(39,32.768)(39,16.384)(39,32.768)(39,16.384)(39,8.192)(39,4.096)(39,2.048)(39,4.096)(38,2.048)(37,1.024)(36,2.048)(35,1.024)(34,2.048)(33,1.024)(32,2.048)(32,1.024)(31,0.512)(30,1.024)(29,0.512)(29,1.024)(28,0.512)(27,0.256)(26,0.512)(25,1.024)(25,0.512)(24,0.256)(23,0.512)(22,0.256)(21,0.512)(21,1.024)(20,0.512)(20,0.256)(19,0.512)(19,1.024)(18,0.512)(18,1.024)(17,2.048)(17,1.024)(17,2.048)(16,1.024)(15,0.512)(15,0.256)(15,0.512)(15,1.024)(15,0.512)(14,0.256)(13,0.512)(13,1.024)(13,2.048)(12,1.024)(12,2.048)(11,4.096)(11,2.048)(11,4.096)(11,2.048)(11,4.096)(11,2.048)(11,4.096)(11,2.048)(11,1.024)(11,2.048)(10,4.096)(10,2.048)(10,1.024)(9,0.512)(9,0.256)(9,0.512)(9,1.024)(8,0.512)(8,1.024)(7,2.048)(6,4.096)(6,2.048)(6,1.024)(6,0.512)(5,0.256)(4,0.128)(3,0.256)(3,0.128)(3,0.256)(3,0.512)(3,0.256)(3,0.128)(3,0.064)(3,0.032)(3,0.064)(3,0.032)(3,0.016)(3,0.008)(2,0.016)(2,0.032)(2,0.064)(2,0.128)(2,0.064)(2,0.032)(2,0.016)(2,0.008)(2,0.004)(2,0.002)(2,0.002)(2,0.004)(2,0.002)(2,0.004)(2,0.008)(2,0.004)(1,0.002)(1,0.004)(1,0.008)(1,0.004)(1,0.002)(1,0.004)(1,0.008)(1,0.004)(1,0.002)(1,0.002)(1,0.002)(1,0.004)(1,0.008)(1,0.016)(1,0.008)(1,0.016)(1,0.032)(1,0.064)(1,0.128)(1,0.064)(1,0.032)(1,0.016)(1,0.032)(1,0.016)(1,0.008)(1,0.016)(1,0.008)(1,0.004)(1,0.002)(1,0.004)};
            \addplot+ coordinates {(485,4.0)(481,8.0)(477,16.0)(468,8.0)(461,16.0)(453,32.0)(446,64.0)(438,32.0)(431,16.0)(426,8.0)(418,4.0)(412,2.0)(408,4.0)(404,8.0)(398,16.0)(392,8.0)(385,16.0)(382,8.0)(376,16.0)(371,32.0)(368,64.0)(367,32.0)(360,64.0)(360,32.0)(357,16.0)(353,32.0)(351,64.0)(351,32.0)(345,16.0)(341,8.0)(339,16.0)(337,8.0)(334,4.0)(331,8.0)(328,16.0)(326,8.0)(324,4.0)(319,8.0)(316,16.0)(314,32.0)(314,16.0)(311,8.0)(309,4.0)(308,2.0)(306,1.0)(305,0.5)(303,1.0)(300,2.0)(298,1.0)(295,0.5)(294,0.25)(292,0.125)(291,0.25)(290,0.125)(289,0.25)(288,0.5)(287,1.0)(285,2.0)(283,1.0)(281,2.0)(278,4.0)(275,2.0)(273,4.0)(271,2.0)(269,4.0)(266,8.0)(264,4.0)(262,2.0)(260,1.0)(259,0.5)(258,1.0)(256,2.0)(255,4.0)(253,2.0)(251,1.0)(250,2.0)(249,4.0)(246,2.0)(244,1.0)(243,0.5)(241,1.0)(239,2.0)(237,4.0)(232,8.0)(230,4.0)(229,2.0)(227,1.0)(226,2.0)(223,1.0)(221,2.0)(218,1.0)(217,0.5)(216,1.0)(214,2.0)(213,4.0)(211,2.0)(210,4.0)(208,2.0)(206,4.0)(205,2.0)(204,1.0)(203,2.0)(201,1.0)(200,0.5)(199,1.0)(197,0.5)(196,1.0)(195,2.0)(193,4.0)(192,2.0)(190,1.0)(189,2.0)(188,1.0)(187,0.5)(186,1.0)(185,0.5)(183,1.0)(182,0.5)(181,0.25)(180,0.5)(179,1.0)(178,2.0)(177,1.0)(176,2.0)(175,1.0)(173,2.0)(171,1.0)(169,0.5)(168,0.25)(167,0.125)(166,0.0625)(165,0.125)(164,0.0625)(163,0.125)(162,0.25)(161,0.125)(160,0.0625)(159,0.03125)(158,0.015625)(157,0.0078125)(156,0.00390625)(155,0.002)(154,0.002)(153,0.002)(152,0.004)(151,0.002)(150,0.002)(149,0.004)(148,0.002)(147,0.004)(146,0.002)(145,0.004)(144,0.002)(143,0.002)(142,0.002)(141,0.002)(140,0.002)(139,0.002)(138,0.002)(137,0.002)(136,0.002)(135,0.002)(134,0.002)(133,0.002)(132,0.002)(131,0.002)(130,0.004)(129,0.002)(128,0.002)(127,0.004)(126,0.008)(125,0.004)(124,0.002)(123,0.004)(122,0.002)(121,0.002)(120,0.002)(119,0.004)(118,0.002)(117,0.004)(116,0.002)(115,0.004)(114,0.008)(113,0.016)(112,0.008)(111,0.016)(110,0.032)(109,0.064)(108,0.128)(107,0.256)(106,0.128)(105,0.064)(104,0.128)(103,0.256)(102,0.128)(101,0.064)(100,0.032)(99,0.016)(98,0.032)(97,0.016)(96,0.032)(95,0.016)(94,0.008)(93,0.004)(92,0.008)(91,0.004)(90,0.002)(89,0.004)(88,0.002)(87,0.002)(86,0.004)(85,0.002)(84,0.002)(83,0.002)(82,0.002)(81,0.002)(80,0.002)(79,0.002)(78,0.004)(77,0.008)(76,0.016)(75,0.008)(74,0.004)(73,0.008)(72,0.016)(71,0.032)(70,0.064)(69,0.032)(68,0.016)(67,0.008)(66,0.016)(65,0.032)(64,0.016)(63,0.008)(62,0.004)(61,0.008)(60,0.004)(59,0.008)(58,0.016)(57,0.008)(56,0.004)(55,0.002)(54,0.004)(53,0.002)(52,0.002)(51,0.002)(50,0.004)(49,0.002)(48,0.002)(47,0.004)(46,0.008)(45,0.004)(44,0.002)(43,0.002)(43,0.004)(42,0.008)(41,0.004)(40,0.002)(39,0.002)(38,0.004)(37,0.002)(36,0.002)(35,0.004)(34,0.002)(33,0.004)(32,0.002)(32,0.004)(32,0.008)(31,0.004)(30,0.008)(29,0.016)(28,0.032)(28,0.016)(27,0.008)(27,0.016)(26,0.008)(25,0.016)(24,0.008)(23,0.016)(23,0.032)(22,0.064)(21,0.128)(20,0.256)(19,0.512)(18,0.256)(18,0.128)(17,0.064)(17,0.128)(16,0.064)(15,0.032)(15,0.064)(14,0.128)(13,0.064)(12,0.128)(12,0.064)(12,0.128)(12,0.064)(12,0.032)(11,0.016)(11,0.008)(10,0.016)(10,0.032)(9,0.016)(8,0.008)(7,0.004)(7,0.002)(6,0.004)(6,0.002)(6,0.002)(6,0.004)(6,0.008)(6,0.016)(5,0.032)(5,0.064)(5,0.032)(5,0.016)(5,0.008)(5,0.016)(4,0.032)(4,0.064)(4,0.032)(4,0.016)(3,0.008)(3,0.004)(2,0.002)(2,0.004)(2,0.002)(2,0.004)(2,0.002)(2,0.002)(2,0.002)(2,0.002)(2,0.004)(2,0.008)(2,0.016)(2,0.032)(2,0.064)(2,0.032)(2,0.064)(2,0.128)(2,0.256)(2,0.128)(2,0.256)(2,0.128)(2,0.064)(2,0.032)(2,0.016)(1,0.008)(1,0.016)(1,0.008)(1,0.004)(1,0.002)(1,0.002)(1,0.004)(1,0.008)(1,0.004)(1,0.008)(1,0.004)(1,0.002)(1,0.004)(1,0.008)(1,0.004)(1,0.002)(1,0.002)(1,0.002)(1,0.002)(1,0.002)(1,0.002)(1,0.002)(1,0.004)(1,0.008)(1,0.016)(1,0.008)(1,0.004)(1,0.002)(1,0.004)(1,0.008)(1,0.004)(1,0.008)(1,0.016)};
        \end{axis}
    \end{tikzpicture} & \begin{tikzpicture}
        \begin{axis}[enlargelimits=false, axis on top, xlabel={Distance to optimum}, width=\mywidth, height=0.22\textheight, ymode=log, point meta min=0, point meta max=1, colorbar, colormap name=viridis, colorbar/width=2.5mm, cycle list name=traces]
            \addplot graphics [xmin=1, xmax=500, ymin=9.765625E-4, ymax=512, includegraphics={trim=0 0 500px 0,clip}]{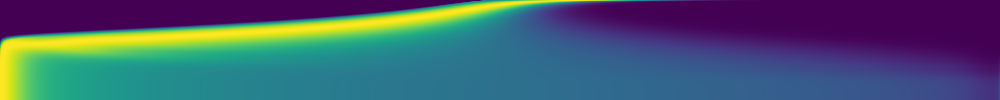}; 
            \addplot+ coordinates {(495,4.0)(486,2.0)(480,4.0)(470,8.0)(457,16.0)(441,32.0)(423,64.0)(405,128.0)(391,64.0)(380,32.0)(369,64.0)(362,32.0)(354,16.0)(341,32.0)(334,64.0)(326,32.0)(319,64.0)(312,32.0)(306,64.0)(302,128.0)(300,250.0)(300,125.0)(300,62.5)(295,31.25)(290,15.625)(285,7.8125)(280,3.90625)(272,2.0)(268,4.0)(263,2.0)(259,4.0)(255,2.0)(251,2.0)(247,4.0)(242,2.0)(239,2.0)(234,4.0)(227,8.0)(223,16.0)(220,8.0)(215,4.0)(212,2.0)(208,2.0)(204,2.0)(202,2.0)(198,4.0)(195,8.0)(190,4.0)(185,8.0)(182,4.0)(180,8.0)(178,16.0)(175,8.0)(172,4.0)(169,8.0)(165,16.0)(162,8.0)(159,16.0)(157,8.0)(155,4.0)(152,2.0)(149,4.0)(146,2.0)(143,2.0)(140,4.0)(137,8.0)(135,16.0)(134,32.0)(134,16.0)(132,32.0)(131,16.0)(129,8.0)(127,4.0)(125,8.0)(122,4.0)(120,2.0)(118,4.0)(116,8.0)(114,4.0)(110,8.0)(107,16.0)(104,8.0)(101,16.0)(99,8.0)(97,4.0)(94,2.0)(91,2.0)(90,4.0)(86,2.0)(83,4.0)(81,2.0)(79,4.0)(77,2.0)(75,2.0)(73,4.0)(71,2.0)(69,2.0)(67,4.0)(65,8.0)(63,4.0)(61,2.0)(59,4.0)(57,2.0)(54,2.0)(52,4.0)(50,8.0)(49,16.0)(49,8.0)(47,4.0)(45,2.0)(43,2.0)(42,4.0)(40,2.0)(38,2.0)(37,2.0)(36,2.0)(35,4.0)(34,2.0)(33,2.0)(32,4.0)(31,8.0)(30,4.0)(29,2.0)(28,2.0)(26,2.0)(24,2.0)(22,2.0)(21,2.0)(20,2.0)(19,4.0)(18,2.0)(17,4.0)(16,2.0)(15,2.0)(14,4.0)(12,2.0)(10,2.0)(9,4.0)(8,2.0)(7,2.0)(5,4.0)(4,8.0)(3,4.0)(2,2.0)(1,2.0)(1,2.0)};
            \addplot+ coordinates {(499,4.0)(490,2.0)(482,2.0)(475,4.0)(466,2.0)(459,4.0)(450,8.0)(441,16.0)(428,32.0)(416,64.0)(404,32.0)(390,64.0)(379,32.0)(369,64.0)(363,32.0)(353,64.0)(344,128.0)(340,64.0)(333,32.0)(326,16.0)(319,8.0)(314,16.0)(310,8.0)(304,16.0)(298,32.0)(294,64.0)(293,32.0)(286,64.0)(281,32.0)(277,64.0)(273,32.0)(268,16.0)(264,8.0)(258,16.0)(252,8.0)(246,16.0)(239,8.0)(234,16.0)(230,8.0)(227,4.0)(221,8.0)(217,4.0)(214,8.0)(211,16.0)(207,8.0)(204,4.0)(200,8.0)(197,4.0)(193,8.0)(190,4.0)(187,2.0)(184,4.0)(180,2.0)(177,4.0)(173,2.0)(170,4.0)(167,8.0)(164,16.0)(162,8.0)(159,16.0)(156,8.0)(154,4.0)(151,2.0)(148,2.0)(144,4.0)(141,8.0)(137,16.0)(135,8.0)(133,16.0)(131,32.0)(131,64.0)(131,32.0)(130,16.0)(128,8.0)(125,4.0)(123,2.0)(121,4.0)(118,8.0)(116,4.0)(113,2.0)(111,2.0)(108,2.0)(105,4.0)(103,8.0)(101,4.0)(99,2.0)(96,4.0)(94,8.0)(92,4.0)(90,2.0)(88,4.0)(86,2.0)(84,4.0)(81,2.0)(78,4.0)(77,8.0)(74,16.0)(74,8.0)(72,4.0)(70,2.0)(68,2.0)(66,4.0)(64,8.0)(62,4.0)(60,2.0)(58,2.0)(57,2.0)(55,2.0)(53,4.0)(51,2.0)(49,2.0)(48,2.0)(46,4.0)(43,2.0)(42,4.0)(40,8.0)(38,4.0)(36,2.0)(35,2.0)(33,4.0)(32,8.0)(31,4.0)(29,2.0)(28,2.0)(26,4.0)(25,2.0)(23,2.0)(21,2.0)(20,4.0)(19,8.0)(18,4.0)(17,2.0)(16,2.0)(15,4.0)(14,8.0)(13,4.0)(12,2.0)(11,4.0)(10,2.0)(9,2.0)(8,2.0)(7,2.0)(6,2.0)(5,2.0)(4,2.0)(3,4.0)(2,2.0)(1,4.0)(1,2.0)};
            \addplot+ coordinates {(465,4.0)(455,2.0)(449,4.0)(439,8.0)(429,4.0)(421,8.0)(411,16.0)(401,32.0)(391,16.0)(379,32.0)(370,64.0)(361,32.0)(354,16.0)(346,32.0)(339,16.0)(328,8.0)(322,4.0)(316,8.0)(309,16.0)(302,32.0)(296,64.0)(292,32.0)(285,64.0)(281,128.0)(281,64.0)(277,32.0)(273,16.0)(268,8.0)(264,16.0)(260,8.0)(255,16.0)(250,32.0)(245,64.0)(245,32.0)(241,16.0)(237,32.0)(233,16.0)(228,32.0)(225,16.0)(221,32.0)(218,16.0)(214,32.0)(211,16.0)(206,32.0)(203,16.0)(199,8.0)(196,4.0)(191,8.0)(187,4.0)(182,8.0)(178,4.0)(175,2.0)(170,4.0)(167,8.0)(164,4.0)(158,8.0)(155,4.0)(152,2.0)(149,2.0)(146,4.0)(143,2.0)(141,2.0)(139,4.0)(136,2.0)(134,4.0)(132,2.0)(130,4.0)(127,8.0)(125,4.0)(123,2.0)(121,4.0)(118,2.0)(115,4.0)(113,2.0)(110,4.0)(108,2.0)(106,2.0)(104,4.0)(99,8.0)(97,16.0)(96,8.0)(93,4.0)(91,2.0)(89,4.0)(85,8.0)(82,4.0)(80,2.0)(78,2.0)(76,4.0)(74,2.0)(72,4.0)(70,2.0)(68,2.0)(66,4.0)(64,2.0)(62,2.0)(60,4.0)(59,2.0)(57,4.0)(55,2.0)(53,4.0)(51,2.0)(50,2.0)(49,2.0)(47,2.0)(46,4.0)(44,2.0)(42,2.0)(40,4.0)(39,2.0)(37,2.0)(36,4.0)(34,2.0)(33,4.0)(32,8.0)(30,16.0)(30,8.0)(28,4.0)(27,2.0)(26,2.0)(25,2.0)(24,2.0)(23,2.0)(22,4.0)(21,8.0)(20,4.0)(19,2.0)(18,2.0)(17,2.0)(16,4.0)(15,8.0)(14,4.0)(13,2.0)(12,4.0)(11,8.0)(11,4.0)(10,2.0)(9,2.0)(8,2.0)(7,4.0)(6,2.0)(5,2.0)(4,2.0)(3,4.0)(2,2.0)(1,4.0)(1,2.0)(1,2.0)};
            \addplot+ coordinates {(475,4.0)(467,8.0)(456,4.0)(448,2.0)(443,4.0)(436,2.0)(430,4.0)(421,8.0)(411,16.0)(401,32.0)(392,16.0)(380,32.0)(368,16.0)(357,32.0)(348,16.0)(340,32.0)(332,16.0)(322,8.0)(316,16.0)(310,8.0)(304,16.0)(298,8.0)(292,16.0)(286,32.0)(278,64.0)(275,32.0)(269,16.0)(264,8.0)(259,16.0)(254,8.0)(249,4.0)(244,2.0)(240,1.0)(237,2.0)(233,4.0)(227,2.0)(222,4.0)(219,2.0)(216,1.0)(213,2.0)(209,4.0)(206,2.0)(202,1.0)(199,0.5)(197,0.25)(194,0.5)(192,1.0)(189,2.0)(186,4.0)(182,8.0)(179,16.0)(176,8.0)(173,4.0)(169,8.0)(166,4.0)(163,2.0)(160,1.0)(157,2.0)(154,4.0)(151,2.0)(148,1.0)(146,2.0)(143,4.0)(140,8.0)(138,4.0)(136,8.0)(133,4.0)(130,8.0)(127,4.0)(125,2.0)(123,4.0)(120,2.0)(117,1.0)(115,0.5)(113,1.0)(111,0.5)(109,1.0)(107,2.0)(105,1.0)(103,2.0)(101,1.0)(98,2.0)(96,1.0)(93,2.0)(91,4.0)(89,2.0)(87,4.0)(84,2.0)(82,4.0)(80,2.0)(78,4.0)(75,2.0)(73,4.0)(71,2.0)(70,1.0)(68,0.5)(66,1.0)(64,2.0)(62,4.0)(60,2.0)(59,1.0)(57,2.0)(55,1.0)(54,2.0)(53,1.0)(51,2.0)(49,1.0)(48,2.0)(46,1.0)(45,0.5)(43,1.0)(42,2.0)(41,4.0)(40,8.0)(39,4.0)(38,8.0)(37,4.0)(36,2.0)(35,4.0)(34,2.0)(33,4.0)(32,2.0)(31,4.0)(29,2.0)(28,4.0)(26,8.0)(24,16.0)(24,8.0)(23,4.0)(22,8.0)(21,4.0)(20,8.0)(19,16.0)(19,32.0)(19,16.0)(19,8.0)(18,4.0)(17,8.0)(16,16.0)(16,8.0)(15,4.0)(14,2.0)(13,1.0)(12,2.0)(11,4.0)(10,8.0)(9,4.0)(8,2.0)(7,1.0)(6,0.5)(5,1.0)(4,0.5)(3,1.0)(2,0.5)(1,1.0)(1,0.5)(1,0.25)};
            \addplot+ coordinates {(490,1.0)(485,0.5)(481,1.0)(476,2.0)(470,4.0)(461,8.0)(451,16.0)(434,32.0)(417,64.0)(400,128.0)(386,64.0)(377,32.0)(367,16.0)(357,32.0)(349,64.0)(343,128.0)(340,64.0)(334,32.0)(327,64.0)(320,32.0)(314,16.0)(307,8.0)(299,16.0)(294,32.0)(288,64.0)(284,32.0)(274,16.0)(268,8.0)(262,16.0)(256,8.0)(250,4.0)(246,2.0)(241,4.0)(236,8.0)(232,4.0)(228,8.0)(224,16.0)(220,8.0)(216,4.0)(211,2.0)(206,1.0)(203,2.0)(199,4.0)(193,8.0)(189,4.0)(185,8.0)(182,4.0)(176,2.0)(173,4.0)(169,2.0)(166,4.0)(164,8.0)(159,16.0)(154,8.0)(151,16.0)(149,8.0)(146,4.0)(143,8.0)(141,4.0)(139,8.0)(136,4.0)(133,2.0)(129,4.0)(125,8.0)(122,4.0)(120,8.0)(118,4.0)(115,2.0)(113,1.0)(110,2.0)(108,4.0)(105,8.0)(102,4.0)(100,8.0)(97,4.0)(95,2.0)(93,1.0)(91,2.0)(89,4.0)(87,8.0)(85,4.0)(83,2.0)(81,1.0)(79,0.5)(77,1.0)(74,2.0)(71,4.0)(69,2.0)(67,1.0)(65,0.5)(63,1.0)(61,2.0)(59,4.0)(57,8.0)(55,4.0)(54,8.0)(53,4.0)(51,2.0)(50,4.0)(49,2.0)(47,4.0)(45,2.0)(44,4.0)(43,8.0)(42,4.0)(40,2.0)(39,1.0)(37,0.5)(35,1.0)(33,0.5)(32,1.0)(31,0.5)(30,0.25)(29,0.125)(28,0.0625)(27,0.125)(26,0.25)(25,0.125)(24,0.0625)(23,0.03125)(22,0.015625)(21,0.03125)(20,0.0625)(19,0.125)(18,0.25)(17,0.125)(16,0.0625)(15,0.03125)(14,0.015625)(13,0.03125)(12,0.015625)(11,0.03125)(10,0.015625)(9,0.03125)(8,0.015625)(7,0.0078125)(6,0.015625)(5,0.0078125)(4,0.00390625)(3,0.0078125)(2,0.015625)(1,0.0078125)};
            \addplot+ coordinates {(465,4.0)(457,8.0)(446,4.0)(435,8.0)(424,16.0)(410,8.0)(400,16.0)(391,32.0)(383,16.0)(373,32.0)(364,16.0)(357,8.0)(349,4.0)(344,2.0)(339,4.0)(332,8.0)(324,16.0)(319,8.0)(314,16.0)(306,8.0)(299,16.0)(292,8.0)(287,16.0)(282,32.0)(276,64.0)(269,32.0)(264,16.0)(255,8.0)(247,4.0)(242,8.0)(235,16.0)(231,32.0)(227,16.0)(221,32.0)(219,16.0)(216,8.0)(213,4.0)(209,8.0)(206,4.0)(203,2.0)(200,4.0)(196,2.0)(193,1.0)(190,0.5)(188,0.25)(186,0.125)(184,0.0625)(183,0.03125)(182,0.015625)(181,0.0078125)(180,0.00390625)(179,0.002)(178,0.002)(177,0.004)(176,0.008)(175,0.016)(174,0.008)(173,0.004)(172,0.008)(171,0.004)(170,0.002)(169,0.002)(168,0.004)(167,0.002)(166,0.004)(165,0.002)(164,0.004)(163,0.002)(162,0.002)(161,0.002)(160,0.002)(159,0.004)(158,0.002)(157,0.002)(156,0.004)(155,0.002)(154,0.002)(153,0.004)(152,0.008)(151,0.004)(150,0.008)(149,0.016)(148,0.008)(147,0.004)(146,0.008)(145,0.016)(144,0.008)(143,0.016)(142,0.032)(140,0.064)(139,0.032)(138,0.016)(137,0.032)(136,0.064)(135,0.128)(133,0.064)(132,0.128)(131,0.256)(130,0.128)(128,0.256)(127,0.128)(126,0.256)(124,0.512)(122,1.024)(120,2.048)(117,4.096)(114,2.048)(111,4.096)(108,2.048)(105,4.096)(103,2.048)(100,1.024)(98,0.512)(96,1.024)(94,2.048)(92,1.024)(90,0.512)(88,1.024)(86,0.512)(84,1.024)(82,0.512)(80,0.256)(78,0.512)(76,1.024)(75,0.512)(73,1.024)(71,2.048)(70,4.096)(68,8.192)(67,16.384)(66,32.768)(66,16.384)(65,8.192)(64,4.096)(63,8.192)(62,4.096)(60,2.048)(58,4.096)(56,8.192)(54,4.096)(52,2.048)(50,1.024)(49,0.512)(48,0.256)(47,0.512)(46,0.256)(45,0.512)(44,0.256)(43,0.128)(42,0.256)(41,0.128)(40,0.256)(39,0.128)(38,0.256)(37,0.512)(36,1.024)(34,2.048)(33,1.024)(32,2.048)(30,1.024)(28,2.048)(27,4.096)(26,2.048)(25,1.024)(24,0.512)(23,1.024)(22,0.512)(21,1.024)(20,2.048)(18,1.024)(17,2.048)(16,1.024)(15,2.048)(14,1.024)(13,2.048)(12,4.096)(11,2.048)(10,1.024)(9,0.512)(8,0.256)(7,0.128)(6,0.064)(5,0.128)(4,0.064)(3,0.128)(2,0.256)(1,0.128)(1,0.064)};
        \end{axis}
    \end{tikzpicture}
    \end{tabular}
		\vspace{-10pt}
    \caption{Parameter control plots of the two-rate method atop parameter efficiency heatmaps, $n=10^3$, $\lambda=64$ (left) and $\lambda=2048$ (right).
             Red traces are for the mutation rate lower bound of $\rho_{\min}=1/n$, black traces are for the lower bound of $\rho_{\min}=1/n^2$}
    \label{fig:performance-vs-heatmaps}
\end{figure}

With such pictures, we can infer how far a certain algorithm is from an optimally tuned algorithm with the same structure, which can highlight its strengths and weaknesses.
However, it is difficult to derive insights from just expected times. To get more information, one can record the parameter values produced by the investigated parameter control method
and draw them atop the heatmaps produced as in Sec.~\ref{sec:sensitivity}. An example of this is shown in Fig.~\ref{fig:performance-vs-heatmaps}. An insight from this figure, that
may be relevant to the analysis of strengths and weaknesses of this method, would be that the version using $\rho_{\min}=1/n$ cannot use very small probabilities
and is thus suboptimal at distances close to the optimum, whereas the version using $\rho_{\min}=1/n^2$ falls down from the optimal parameter region too frequently and too deep.

\section{Conclusions}
\label{sec:cond}

Extending the work~\cite{BuskulicD19}, we have presented in this work optimal and drift-maximizing mutation rates for two different $(1+\lambda)$~EAs (using standard bit mutation and shift mutation, respectively) 
and for the $(1+\lambda)$~RLS. 
We have demonstrated how our data can be used to detect weak spots of parameter control mechanisms.
We have also described two unexpected effects of the dependency of the expected remaining optimization time on the mutation rates: non-monotonicity in $d$ (Sec.~\ref{sec:opt}) and non-unimodality (Sec.~\ref{sec:sensitivity}). We plan on exploring these effects in more detail, and with mathematical rigor. Likewise, we plan on analyzing the formal relationship of the optimal mutation rates 
with the normalized distance $d/n$. 
As a first step towards this goal, we will use the numerical data presented above to derive close-form expressions for the expected remaining optimization times $T_{D,O}(n,\lambda,d,\rho)$ as well as for the optimal configurations $\rho^{*}_{D,O}(n,\lambda,d)$. 
Finally, we also plan on applying similar analyses to more sophisticated benchmark problems.

%
%

\textbf{Acknowledgments.} The work was supported by RFBR and CNRS, project no. 20-51-15009, by the Paris Ile-de-France Region, and by ANR-11-LABX-0056-LMH.

\renewcommand{\doi}[1]{}
\bibliographystyle{splncs04}
\bibliography{../../../../bibliography,../carola}

\newpage
\begin{landscape}
\appendix
\section{Derivation of the Drift Values in Table~\ref{tbl:example-drift}}
\label{sec:example-drift-derivation}

We recall that for \OM the probability $P(d,d',\rho)$ to sample one individual at distance $d' < d$ from the optimum
when the parent is at distance $d$ from the optimum by flipping exactly $\rho$ bits equals
\begin{align*}
\end{align*}
For convenience we say that $P(d,d,\rho) = 1 - \sum_{t=0}^{d-1} P(d,t,\rho)$, which has the meaning under elitist selection.
Under elitist selection, it holds that $P(d,d,\rho)$ can be expressed as $1 - \sum_{t=0}^{d-1} P(d,t,\rho)$.

The values for $n=30$, $d=7$, $d'\in[0..7]$, $\rho\in[1..10]$ are presented below.

\vspace{1ex}
\begingroup\centering\scriptsize
\begin{tabular}{c|*{10}{c}}
$d'$ & 1 & 2 & 3 & 4 & 5 & 6 & 7 & 8 & 9 & 10 \\\hline
0 & 0 & 0 & 0 & 0 & 0 & 0 & $\frac{1}{2035800}$ & 0 & 0 & 0\\
1 & 0 & 0 & 0 & 0 & 0 & $\frac{1}{84825}$ & 0 & $\frac{1}{254475}$ & 0 & 0\\
2 & 0 & 0 & 0 & 0 & $\frac{1}{6786}$ & 0 & $\frac{161}{2035800}$ & 0 & $\frac{1}{56550}$ & 0\\
3 & 0 & 0 & 0 & $\frac{1}{783}$ & 0 & $\frac{23}{28275}$ & 0 & $\frac{77}{254475}$ & 0 & $\frac{1}{16965}$\\
4 & 0 & 0 & $\frac{1}{116}$ & 0 & $\frac{115}{20358}$ & 0 & $\frac{1771}{678600}$ & 0 & $\frac{49}{56550}$ & 0\\
5 & 0 & $\frac{7}{145}$ & 0 & $\frac{23}{783}$ & 0 & $\frac{253}{16965}$ & 0 & $\frac{539}{84825}$ & 0 & $\frac{7}{3393}$\\
6 & $\frac{7}{30}$ & 0 & $\frac{69}{580}$ & 0 & $\frac{1265}{20358}$ & 0 & $\frac{12397}{407160}$ & 0 & $\frac{49}{3770}$ & 0\\
\hline
7 & $\frac{23}{30}$ & $\frac{138}{145}$ & $\frac{253}{290}$ & $\frac{253}{261}$ & $\frac{6325}{6786}$ & $\frac{5566}{5655}$ & $\frac{98417}{101790}$ & $\frac{16852}{16965}$ & $\frac{11153}{11310}$ & $\frac{1881}{1885}$
\end{tabular}
\par\endgroup
\vspace{2ex}

The values for $n=30$, $d=8$, $d'\in[0..8]$, $\rho\in[1..10]$ are presented below.

\vspace{1ex}
\begingroup\centering\scriptsize
\begin{tabular}{c|*{10}{c}}
$d'$ & 1 & 2 & 3 & 4 & 5 & 6 & 7 & 8 & 9 & 10 \\\hline
0 & 0 & 0 & 0 & 0 & 0 & 0 & 0 & $\frac{1}{5852925}$ & 0 & 0\\
1 & 0 & 0 & 0 & 0 & 0 & 0 & $\frac{1}{254475}$ & 0 & $\frac{1}{650325}$ & 0\\
2 & 0 & 0 & 0 & 0 & 0 & $\frac{4}{84825}$ & 0 & $\frac{176}{5852925}$ & 0 & $\frac{1}{130065}$\\
3 & 0 & 0 & 0 & 0 & $\frac{4}{10179}$ & 0 & $\frac{77}{254475}$ & 0 & $\frac{28}{216775}$ & 0\\
4 & 0 & 0 & 0 & $\frac{2}{783}$ & 0 & $\frac{176}{84825}$ & 0 & $\frac{2156}{1950975}$ & 0 & $\frac{32}{78039}$\\
5 & 0 & 0 & $\frac{2}{145}$ & 0 & $\frac{110}{10179}$ & 0 & $\frac{539}{84825}$ & 0 & $\frac{392}{130065}$ & 0\\
6 & 0 & $\frac{28}{435}$ & 0 & $\frac{176}{3915}$ & 0 & $\frac{154}{5655}$ & 0 & $\frac{17248}{1170585}$ & 0 & $\frac{532}{78039}$\\
7 & $\frac{4}{15}$ & 0 & $\frac{22}{145}$ & 0 & $\frac{308}{3393}$ & 0 & $\frac{539}{10179}$ & 0 & $\frac{3724}{130065}$ & 0\\
\hline
8 & $\frac{11}{15}$ & $\frac{407}{435}$ & $\frac{121}{145}$ & $\frac{1243}{1305}$ & $\frac{3047}{3393}$ & $\frac{5489}{5655}$ & $\frac{47861}{50895}$ & $\frac{88616}{90045}$ & $\frac{125932}{130065}$ & $\frac{129124}{130065}$
\end{tabular}
\par\endgroup
\vspace{2ex}

\newpage
We also recall that the probability $P^{\lambda}(d, d', \rho)$ for the best of $\lambda$ offspring to appear exactly at distance $d'$ when the parent is at distance $d$
and exactly $\rho$ bits are flipped in each offspring is:
\begin{align*}
P^{\lambda}(d, d', \rho) &= \left(\sum_{t=d'}^{d} P(d,t,\rho)\right)^{\lambda} - \left(\sum_{t=d'+1}^{d} P(d,t,\rho)\right)^{\lambda}.
\end{align*}

The values for $n=30$, $\lambda=512$, $d=7$, $d'\in[0..7]$, $\rho\in[1..10]$ are presented below, trying best to preserve precision visually.

\vspace{1ex}
\begingroup\centering\scriptsize
\begin{tabular}{c|*{10}{c}}
$d'$ & 1 & 2 & 3 & 4 & 5 & 6 & 7 & 8 & 9 & 10 \\\hline
0 & 0 & 0 & 0 & 0 & 0 & 0 & $2.51\cdot 10^{-4}$ & 0 & 0 & 0\\
1 & 0 & 0 & 0 & 0 & 0 & $6.02\cdot 10^{-3}$ & 0 & $2.01\cdot 10^{-3}$ & 0 & 0\\
2 & 0 & 0 & 0 & 0 & $7.27\cdot 10^{-2}$ & 0 & $3.97\cdot 10^{-2}$ & 0 & $9.01\cdot 10^{-3}$ & 0\\
3 & 0 & 0 & 0 & $4.80\cdot 10^{-1}$ & 0 & $3.39\cdot 10^{-1}$ & 0 & $1.43\cdot 10^{-1}$ & 0 & $2.97\cdot 10^{-2}$\\
4 & 0 & 0 & $1 - 1.19\cdot 10^{-2}$ & 0 & $1 - 1.24\cdot 10^{-1}$ & 0 & $1 - 2.92\cdot 10^{-1}$ & 0 & $3.55\cdot 10^{-1}$ & 0\\
5 & 0 & $1 - 9.95\cdot 10^{-12}$ & 0 & $1 - 4.80\cdot 10^{-1}$ & 0 & $1 - 3.45\cdot 10^{-1}$ & 0 & $1 - 1.78\cdot 10^{-1}$ & 0 & $1 - 3.67\cdot 10^{-1}$\\
6 & $1 - 8.29\cdot 10^{-60}$ & 0 & $1.19\cdot 10^{-2}$ & 0 & $5.10\cdot 10^{-2}$ & 0 & $2.52\cdot 10^{-1}$ & 0 & $1 - 3.65\cdot 10^{-1}$ & 0\\
\hline
7 & $8.29\cdot 10^{-60}$ & $9.95\cdot 10^{-12}$ & $4.47\cdot 10^{-31}$ & $1.20\cdot 10^{-7}$ & $2.27\cdot 10^{-16}$ & $2.97\cdot 10^{-4}$ & $3.21\cdot 10^{-8}$ & $3.27\cdot 10^{-2}$ & $7.79\cdot 10^{-4}$ & $3.37\cdot 10^{-1}$\\
\end{tabular}
\par\endgroup
\vspace{2ex}

The values for $n=30$, $\lambda=512$, $d=8$, $d'\in[0..8]$, $\rho\in[1..10]$ are presented below, trying best to preserve precision visually.

\vspace{1ex}
\begingroup\centering\scriptsize
\begin{tabular}{c|*{10}{c}}
$d'$ & 1 & 2 & 3 & 4 & 5 & 6 & 7 & 8 & 9 & 10 \\\hline
0 & 0 & 0 & 0 & 0 & 0 & 0 & 0 & $8.75\cdot 10^{-5}$ & 0 & 0\\
1 & 0 & 0 & 0 & 0 & 0 & 0 & $2.01\cdot 10^{-3}$ & 0 & $7.87\cdot 10^{-4}$ & 0\\
2 & 0 & 0 & 0 & 0 & 0 & $2.39\cdot 10^{-2}$ & 0 & $1.53\cdot 10^{-2}$ & 0 & $3.93\cdot 10^{-3}$\\
3 & 0 & 0 & 0 & 0 & $1.82\cdot 10^{-1}$ & 0 & $1.43\cdot 10^{-1}$ & 0 & $6.39\cdot 10^{-2}$ & 0\\
4 & 0 & 0 & 0 & $1 - 2.70\cdot 10^{-1}$ & 0 & $1 - 3.61\cdot 10^{-1}$ & 0 & $4.26\cdot 10^{-1}$ & 0 & $1.89\cdot 10^{-1}$\\
5 & 0 & 0 & $1 - 8.16\cdot 10^{-4}$ & 0 & $1 - 1.85\cdot 10^{-1}$ & 0 & $1 - 1.78\cdot 10^{-1}$ & 0 & $1 - 2.64\cdot 10^{-1}$ & 0\\
6 & 0 & $1 - 1.61\cdot 10^{-15}$ & 0 & $2.70\cdot 10^{-1}$ & 0 & $3.37\cdot 10^{-1}$ & 0 & $1 - 4.41\cdot 10^{-1}$ & 0 & $1 - 2.17\cdot 10^{-1}$\\
7 & $1 - 1.08\cdot 10^{-69}$ & 0 & $8.16\cdot 10^{-4}$ & 0 & $3.13\cdot 10^{-3}$ & 0 & $3.27\cdot 10^{-2}$ & 0 & $1.99\cdot 10^{-1}$ & 0\\
\hline
8 & $1.08\cdot 10^{-69}$ & $1.61\cdot 10^{-15}$ & $5.83\cdot 10^{-41}$ & $1.50\cdot 10^{-11}$ & $1.21\cdot 10^{-24}$ & $2.37\cdot 10^{-7}$ & $2.15\cdot 10^{-14}$ & $2.77\cdot 10^{-4}$ & $6.60\cdot 10^{-8}$ & $2.43\cdot 10^{-2}$\\
\end{tabular}
\par\endgroup
\vspace{2ex}

Finally, the drift for $\rho$ is $\sum_{d'=0}^{d-1} (d-d') \cdot P^{\lambda}(d, d',\rho)$, so the drift values are, for both $d$, with row-best values bold:

\vspace{1ex}
\begingroup\centering\scriptsize
\begin{tabular}{c|*{10}{c}}
$d$ & 1 & 2 & 3 & 4 & 5 & 6 & 7 & 8 & 9 & 10 \\\hline
7 & 0.5000 & 2.0000 & 2.9762 & 2.9604 & \textbf{3.0434} & 2.7009 & 2.5766 & 2.2292 & 1.7457 & 1.3854\\
8 & 0.5000 & 2.0000 & 2.9984 & \textbf{3.4601} & 3.3583 & 3.3737 & 3.2292 & 2.9124 & 2.7323 & 2.3445\\
\end{tabular}
\par\endgroup
\vspace{2ex}
\end{landscape}

\end{document}